\documentclass{article}
\usepackage[final]{corl_2020} 

\usepackage[utf8]{inputenc}
\usepackage{graphicx}
\usepackage{wrapfig}
\usepackage[font=small]{caption}
\usepackage{bm}
\usepackage{amsmath}
\usepackage{amssymb}
\usepackage{bbm}
\usepackage{url}
\usepackage{mathrsfs}
\usepackage[table]{xcolor}
\usepackage{color}
\usepackage{dsfont}
\usepackage{xspace}
\usepackage{listings}
\usepackage{numprint}
\usepackage{multirow}
\usepackage{siunitx}
\usepackage{setspace,lipsum}
\usepackage{microtype}
\usepackage{subcaption}
\usepackage{booktabs} 
\usepackage[backgroundcolor=white]{todonotes}
\usepackage{overpic}
\usepackage{mathtools}
\usepackage{placeins}
\usepackage{makecell}

\newcommand{\norm}[1]{\left\lVert#1\right\rVert}
\newcommand{\mcL}{\mathcal{L}}

\makeatletter
\DeclareRobustCommand\onedot{\futurelet\@let@token\@onedot}
\def\@onedot{\ifx\@let@token.\else.\null\fi\xspace}

\def\eg{\emph{e.g}\onedot} 
\def\ie{\emph{i.e}\onedot}

\makeatother

\newcommand\blfootnote[1]{%
  \begingroup
  \renewcommand\thefootnote{}\footnote{#1}%
  \addtocounter{footnote}{-1}%
  \endgroup
}

\title{Learning Object Manipulation Skills via Approximate State Estimation from Real Videos}

\author{
Vladim\'ir Petr\'ik$^{1*}$, \hspace{2mm}
Makarand Tapaswi$^{2*}$, \hspace{2mm}
Ivan Laptev$^{2}$, \hspace{2mm}
Josef \v{S}ivic$^{1}$ \\
$^1$CIIRC, Czech Technical University in Prague \hspace{4mm} $^2$Inria \\
{\small \texttt{\{vladimir.petrik,josef.sivic\}@cvut.cz, \{makarand.tapaswi,ivan.laptev\}@inria.fr}}
}

\begin{document}

\maketitle
\blfootnote{*indicates equal contribution}
\vspace{-5mm}
\begin{abstract}
Humans are adept at learning new tasks by watching a few instructional videos.
On the other hand, robots that learn new actions either require a lot of effort through trial and error, or use expert demonstrations that are challenging to obtain.
In this paper, we explore a method that facilitates learning object manipulation skills directly from videos.
Leveraging recent advances in 2D visual recognition and differentiable rendering, we develop an optimization based method to estimate a coarse 3D state representation for the hand and the manipulated object(s) without requiring any supervision.
We use these trajectories as dense rewards for an agent that learns to mimic them through reinforcement learning.
We evaluate our method on simple single- and two-object actions from the Something-Something dataset.
Our approach allows an agent to learn actions from single videos, while watching multiple demonstrations makes the policy more robust.
We show that policies learned in a simulated environment can be easily transferred to a real robot.
\end{abstract}

\vspace{-3mm}

\section{Introduction}
\label{sec:intro}

\begin{wrapfigure}{r}{6cm}
\vspace{-4mm}
\centering
\includegraphics[width=\linewidth]{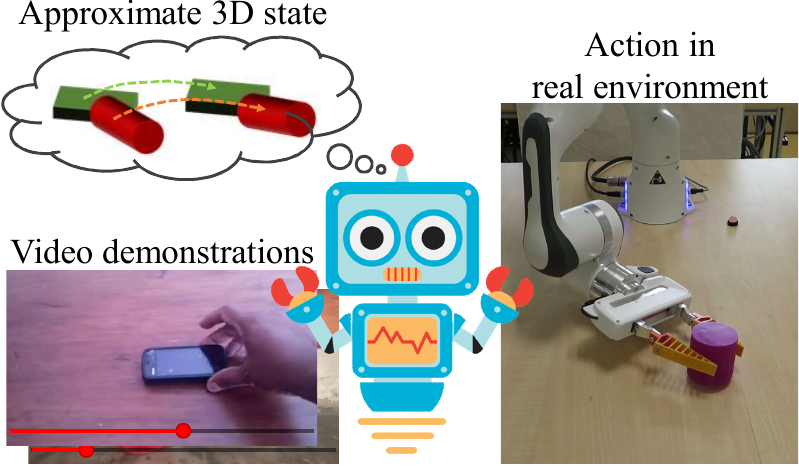}
\vspace{-4mm}
\caption{Our robot watches a few video demonstrations and learns to perform the observed action by estimating a coarse hand-object 3D state.}
\label{fig:teaser}
\vspace{-4mm}
\end{wrapfigure}

Imagine that you want to assemble your bicycle.
Humans are able to watch and learn from other people that demonstrate how to perform these actions, \eg, by watching videos from YouTube~\cite{miech2019howto100m}.
We are interested in providing intelligent agents the ability to learn object manipulation skills by watching videos (see Fig.~\ref{fig:teaser}).

Learning from demonstrations~\cite{argall2009survey} aims at teaching robots to perform various actions based on demonstrations, often performed by \emph{experts}.
While previous works in this area have required humans to teleoperate the robot~\cite{zhang2018imitationComplexTasks}, we wish to address a setting where the robot learns directly from video demonstrations.
In particular, we are interested in understanding how to bridge the gap between the observed moving pixels (videos) and instructions that a robot can understand and execute.
We address this question in two steps (see Fig.~\ref{fig:model}).
First, building upon advances in computer vision and differentiable rendering, we propose \emph{Real2Sim}, a method that lifts real world 2D videos to an approximate 3D state space representation of the hand and the manipulated objects (Sec.~\ref{sec:real2sim}).
Second, we use these automatically extracted trajectories along with reinforcement learning (RL) to learn policies that execute the actions in a 3D simulation environment corresponding to the real robot set-up and on a real robot (Sec.~\ref{sec:rl}).

While there have been recent works on estimating detailed 3D meshes for hands and manipulated objects~\cite{hasson2019obman}, these methods do not generalize well to out-of-domain videos.
Instead, we pursue an alternative solution: we extract only a coarse representation of the scene where we approximate the hand as a cylinder and objects as cuboids.
Leveraging 2D hand-object detectors~\cite{shan2020handobject} and pixel-segmentation methods~\cite{he2017maskrcnn}, we are able to reconstruct a coarse 3D state representation that shares perceptual similarities with the video.
Even though the estimated states lack details (\eg~grasping an object), the coarse states of the dynamic scene act as a guide (a dense reward) for RL in a simulated environment, that in turn resembles the real world sufficiently allowing to transfer the learned policy to the real world.
Our intermediate state representation allows a tangible and interpretable parsing of the video, and is specially effective at modeling actions that include object motion.

We investigate the following three questions in this work.
First, we are interested in studying whether we can learn to perform simple object manipulations based on a single demonstration.
We evaluate our approach on 9 simple object manipulation actions from the Something-Something action recognition dataset~\cite{goyal2017something} and show that some \emph{clean} videos are indeed capable of teaching an agent how to perform the task.
Second, we hypothesize and empirically show that a robust policy can be trained by watching a handful of videos (a few-shot setting), thus reducing the impact of variable quality of individual demonstrations (Sec.~\ref{sec:evaluation}).
Finally, we are interested in studying how the object size or the initial gripper position impacts performance.
We show that automatic domain randomization~\cite{akkaya2019solving} presents a form of curriculum learning strategy that allows the agent to learn the task with gradually increasing complexity.
We propose a challenging benchmark over the 9 actions, with randomized object sizes and initial gripper positions.
Our action-specific metrics analyze whether the robot is able to perform the action correctly.
We also demonstrate how the learned policies can be transferred to a real robot.
We will make the code and data publicly available at \url{https://git.io/JTPkj}.

\section{Related Work}
\label{sec:relwork}

We discuss recent advances in extracting 3D state/mesh representations from 2D images or videos and learning from video demonstrations.


\textbf{Differentiable rendering.}
Understanding the 3D world in a projected image is a classical computer vision problem~\cite{roberts1963blocksworld,gupta2010blocksworld}.
However, obtaining ground-truth labels for 3D meshes is much harder than 2D object bounding boxes leading to challenges in 3D modeling~\cite{gkioxari2019meshrcnn}.
One approach is to learn a \emph{de-renderer}~\cite{wu2017nsd}, an encoder that attempts to predict a structured scene representation, which can reconstruct the original scene through a graphics engine.
This is extended to study object dynamics through visual de-animation~\cite{wu2017vsd}.
Alternatively, there have been attempts to reconstruct 3D object meshes by incorporating graph neural networks~\cite{wang2018pixel2mesh} or intermediate voxel spaces~\cite{gkioxari2019meshrcnn}.
Recent developments in differentiable neural rendering~\cite{kato2018renderer,liu2019softrasterizer}, have enabled approaches that estimate both the 3D texture and shape~\cite{chen2019dib} while only requiring 2D supervision (\eg~a segmentation mask).
Ours is a parameter-free method that can estimate an approximate 3D representation from a single video.

\textbf{Hands and objects in 3D.}
Joint understanding of hands and objects has implications from action recognition~\cite{damen2020epickitchens,tekin2019hoact} to virtual or agumented reality~\cite{holl2018vrhand,tang2020arhands}.
Inspired by the SMPL person model~\cite{bogo2016smpl}, \citet{romero2017mano} develop MANO, an articulated hand model has helped researchers focus on studying various hand deformations~\cite{zhou2020monocular}.
Recently, there have been large data collection efforts for the joint study of hands and objects: as 2D bounding boxes~\cite{shan2020handobject,materzynska2020sthelse}, or 3D pose estimates~\cite{hampali2020honnotate}.
Nevertheless, joint 3D reconstruction of hand and object, especially during manipulation remains a challenging problem.
There have been some efforts in this direction, such as leveraging synthetic datasets and utilizing manipulation constraints~\cite{hasson2019obman}, or addressing the challenges in 3D annotation through temporal consistency in a video~\cite{hasson2020handobjecttime}.
In this work, we propose an alternative solution.
We hypothesize that a coarse 3D state representation is sufficient to teach a robot to perform these actions, and propose an optimization procedure to estimate hand-object trajectories from monocular videos.

\textbf{Learning from demonstrations.}
Enabling a robot to learn policies based on expert demonstrations is a well-studied problem~\cite{argall2009survey,hazara2016contact,vecerik2017demonstrations}.
Learning from demonstrations (LfD) circumvents designing task-specific reward functions~\cite{yu2019metaworld} that are often challenging to devise.
Many works in this direction leverage humans teleoperating a robot, \eg~\cite{zhang2018imitationComplexTasks,caccavale2019kinesthetic,strudel2020rlbc}.
In contrast, we are interested in leveraging the large diversity of videos where people perform various object manipulations
as our demonstrations.

\textbf{Learning from videos}
include directions such as estimating perceptual reward functions from a small number of demonstrations~\cite{sermanet2017perceptualrewards};
learning a visual representation via multiple views and metric learning, followed by learning a policy using a single third person demonstration~\cite{sermanet2017tcn};
imitation from observation by learning to predict different viewpoints~\cite{liu2018imitationObservation};
or very recently, learning pixel-level translation to convert human demonstration to the robot's perspective~\cite{smith2020avid}.
Our goal is similar, but our approach differs.
Without the need for multiple views, we estimate a 3D physical state representation through an optimization algorithm, thus mapping the video to a simulated environment.
Parallel to our work, \citet{shao2020concept2robot} present Concept2Robot that leverages an action classifier as a reward signal.
They learn a single multi-task policy with a natural language instruction encoder that generalizes to small variations in actions.
Nevertheless, the limited performance of the classifier, especially on simulated videos leads to noisy reward signals.

\textbf{Mimicking actions.}
3D human reconstruction~\cite{kanazawa2018hmr} has seen use in teaching robots to mimic human actions~\cite{peng2018deepMimic,peng2018sfv,peng2020imitatingAnimals}.
While human pose can be extracted from a video with surprising accuracy, this is currently not the case for estimating the pose of 3D objects manipulated by hands.
Our approximate scene representation addresses these challenges as it consists of coarse 3D blocks representing the main objects (and the hand) in the input video.
In addition, our videos/demonstrations are aligned automatically by leveraging components of the state representation.


\section{\emph{Real2Sim}: Approximate State Estimation from Video}
\label{sec:real2sim}

We model the structure and motion of the hand and object(s) in the video with coarse 3D models: a hand (wrist onwards) is represented as a cylinder with a fixed height and radius approximating the dimensions of a real hand, while objects are approximated as cuboids.
In this section, we present our approach for 2D spatio-temporal video parsing, define the state space, and design perceptual and physics-based losses that optimize and estimate the states for a video.

\textbf{Video pre-processing.}
We parse the video demonstrations using 2D visual reasoning tools:
Mask-RCNN~\cite{he2017maskrcnn} for segmentation masks,
and Hand-Object detector~\cite{shan2020handobject} that also provides a binary output to indicate touching or not.
To ablate detection errors, we use ground-truth boxes by~\cite{materzynska2020sthelse}.
The final outputs of this module, obtained through a Kalman filter, include frame-level bounding boxes and segmentation masks for the hand and object(s).
Please refer to the Appendix~\ref{sec:app:details:preproc} for details.

\begin{figure}[t]
\centering
\includegraphics[width=\linewidth]{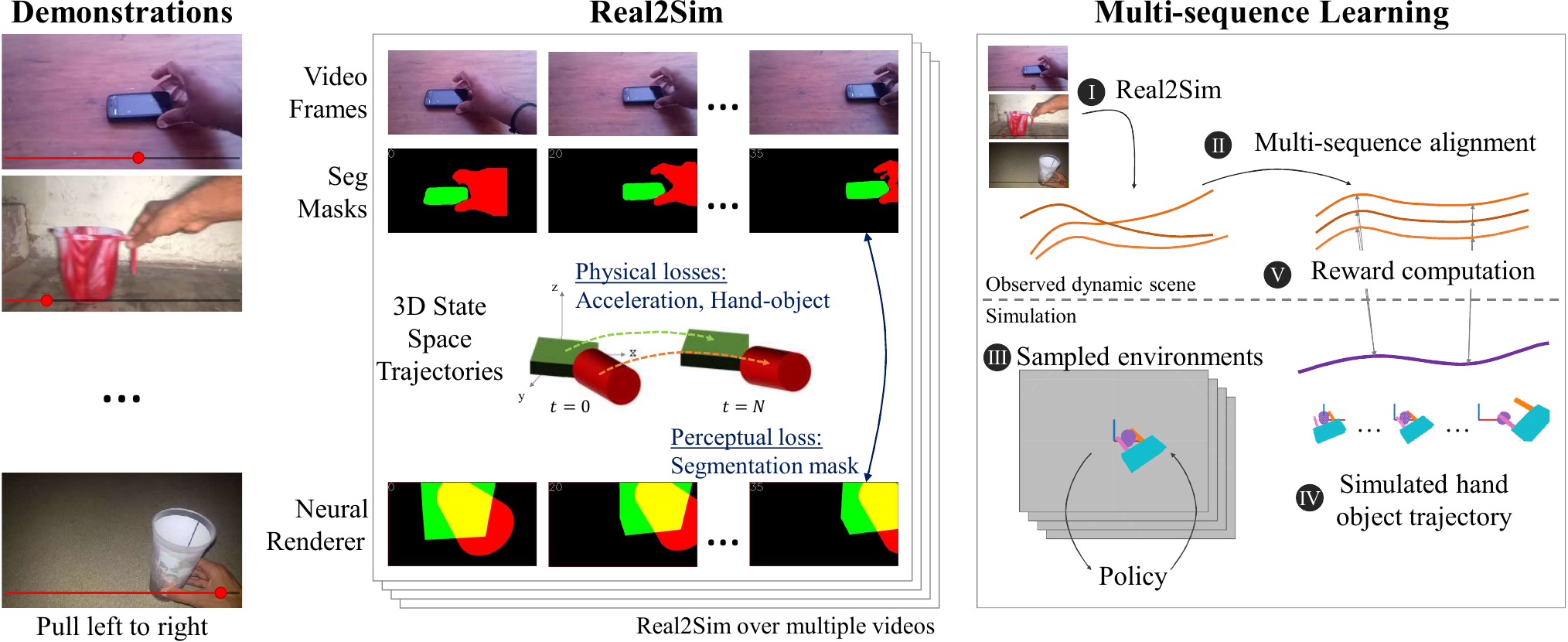}
\caption{Overview of our method.
\textbf{Demonstrations} (left): Our input is a few video demonstrations for each action, here depicting \emph{pull left to right}.
\textbf{Real2Sim} (center): Our optimization based approach obtains coarse 3D state representations for the hand (cylinder) and object (cuboid) based on a combination of physical losses modeling hand-object interactions and perceptual losses measuring similarity between the rendered image and a segmentation mask.
\textbf{Multi-sequence learning} (right):
I. Our approach estimates several trajectories, one for each video.
II. We align these 3D trajectories in space and time.
III. Environments with random object size and starting hand position are sampled.
IV. The learning policy generates a simulated hand-object trajectory.
V. We compute a dense reward that trains the agent to mimic the estimated trajectories.
}
\vspace{-3mm}
\label{fig:model}
\end{figure}

\subsection{Approximate 3D state space}
\label{subsec:real2sim:states}

Consider an input video $\bm v = (v_1, \ldots, v_T)$ with $T$ frames, demonstrating how to perform some action, \eg~\emph{pull left to right} (see Fig.~\ref{fig:model}).
We define a state space to model the hand-object interaction, that mimics the visual content.
(i) We assume the camera is fixed for each video and localize it in 3D using distance to origin, azimuth, and elevation.
(ii) The hand is modeled as a cylinder (\SI{40}{\milli\metre} radius and \SI{150}{\milli\metre} height), with one of the edges symbolizing the manipulator/fingers.
We use a 5D representation: $\bm h = (\bm h_p, \bm h_\theta)$, where $\bm h_p$ encodes the 3D position in Cartesian space and $\bm h_\theta$ captures two rotations: azimuth and elevation.
(iii) Each object, represented as cuboid, is encoded by a 9D vector $\bm o = (\bm o_p, \bm o_s, \bm o_\theta)$ corresponding to 3D position in Cartesian coordinates $\bm o_p$, 3D object size $\bm o_s$, and the angle-axis formulation for the object rotation $\bm o_\theta$.
(iv) Finally, we also encode whether the hand is touching the object as a binary label $\bm \tau$ for each frame.
This leads to a $D = 18$ dimensional state representation (camera: 3, hand: 5, object: 9, touch: 1) for videos/actions with \emph{one} object.

\subsection{Losses and optimization}
\label{subsec:real2sim:losses}
Our goal is to estimate a 3D state trajectory in $\mathbb{R}^{T \times D}$ over the $T$ video frames that mimics the visual content of the demonstration.
We employ two categories of loss functions, \emph{perceptual} and \emph{physical}.
While the general problem of 2D to 3D has multiple solutions, two factors work in our favor:
a fixed hand size helps estimate the distance between the camera and the hand; and a non-zero camera elevation that mimics ego-views helps estimates the depth of the objects placed on the surface.

\noindent \textbf{Perceptual loss}
aims to estimate a 3D state trajectory that closely resembles, when projected, the video.
We achieve this by representing the hand (cylinder) and objects (cuboids) as triangular meshes~\cite{trimesh} that are rendered through a differentiable neural renderer~\cite{kato2018renderer}, projecting the 3D state space of the hand and object to an image.
We render binary object silhouettes without textures as they can be easily compared against the binary segmentation masks extracted from the video.
As the optimization uses predicted segmentation masks, it can be considered as a self-supervised approach.
The perceptual loss measures the reprojection error summed over all frames of the video for the hand and object(s) of interest in the scene:
\begin{equation}
L_{\mathrm{perceptual}} = \sum\nolimits_{t=1}^T \left[ \mathbbm{1}(m^h_t) \cdot \| r^h_t - m^h_t \|^2 + \sum\nolimits_i \mathbbm{1}(m^{o_i}_t) \cdot \| r^{o_i}_t - m^{o_i}_t \|^2 \right] \, ,
\end{equation}
where $m^h_t, m^{o_i}_t$ correspond to the predicted hand and $i$th object segmentation masks and $r^h_t, r^{o_i}_t$ correspond to rendered outputs.
As frame segmentation or tracking may be unreliable, we make the estimation robust by applying the loss only when the mask is non-zero, indicated as $\mathbbm{1}(\cdot)$.
Additionally, a few missed detections are also interpolated using physics-based losses described next.

\noindent \textbf{Physics}
based losses provide regularization and model the hand-object interaction.
As part of the regularization, we minimize the acceleration (double-derivative in time) of all the object positions $\ddot{\bm o}_p$ and rotations $\ddot{\bm o}_\theta$, and the hand position $\ddot{\bm h}_p$ and rotation $\ddot{\bm h}_\theta$:
\begin{equation}
L_{\mathrm{acc}} = \| \ddot{\bm h}_p \| + \| \ddot{\bm h}_\theta \| + \sum\nolimits_i \| \ddot{\bm o}^i_p \| + \| \ddot{\bm o}^i_\theta \| \, .
\end{equation}
Acceleration loss smooths the trajectory and approximates the law of conservation of momentum.
Additionally, as the object is assumed to be non-deformable, we also minimize the distance between the object size $\bm o_s$ and its mean over time $\bar{\bm o}_s$, $L_{\mathrm{size}} = \sum_i \sum_t \| \bm o^i_s(t) - \bar{\bm o}^i_s \|$.

Finally, we regularize the hand-object interaction by imitating the law of inertia with infinite friction.
We assume that only one object is manipulated at any time, and encourage the model to position the hand to be closer to the object if it is moving, and to minimize the object velocity otherwise:
\begin{equation}
L_{\mathrm{interact}} = \bm p_{ho} \cdot \| \dot{\bm o}_p - \dot{\bm h}_p \| + (1 - \bm p_{ho}) \cdot \| \dot{\bm o}_p \| \, ,
\end{equation}
where $\bm p_{ho} = 1-\sigma(\| \bm h_p - \bm o_p \|)$ is the probability that the hand is touching an object based on their positions, and $\sigma()$ is a parameterized sigmoid function that takes in a positive distance value and produces outputs in the range $[0, 1]$.
The first term ensures that the hand and object move together when the hand touches the object and the second term penalizes object motion when the hand is not touching the object.
Hand-object interactions are also encoded by the touch indicator $\bm \tau$ predicted by~\cite{shan2020handobject}.
For the frames where the hand is said to touch the object, we set $\bm h_p = \bm o_p$.

\noindent \textbf{Optimization.}
We use gradient descent to estimate the 3D state space trajectory that minimizes the total loss, a weighted combination of all terms:
$\mcL = w_p L_{\mathrm{perceptual}} + w_a L_{\mathrm{acc}} + w_s L_{\mathrm{size}} + w_i L_{\mathrm{interact}}$.
Owing to the differential neural renderer, we are able to backpropagate through the perceptual loss, while all other terms use standard differentiable components.
See Appendix~\ref{sec:app:details:real2sim} for details.

\section{Learning Object Manipulation Policy from Multiple Videos}
\label{sec:rl}

The objective of RL is to learn to manipulate object(s) so as to imitate the trajectories extracted from the videos.
We use these trajectories to construct a reward function that trains the policy in a physics-based simulator mimicking the real world.
To learn from multiple videos, trajectories are first spatio-temporally aligned.
Note that, \emph{across all actions, we use the same reward function} that encourages the policy to learn to mimic hand/object trajectories.
This is a key aspect of our work as it does not require handcrafting a new reward for each action.
Fig.~\ref{fig:model} (right) presents an overview.


\subsection{Spatio-temporal alignment of multiple extracted trajectories}

Prior to using the trajectories from multiple demonstrations as part of the reward, we spatially align all trajectories of the same action by compensating for the camera orientation and the initial object position.
After the alignment, all trajectories are expressed in the same reference frame given by the starting object position.
In the case of multi-object actions we used the non-manipulated object for the alignment (see Fig.~\ref{fig:alignment} in Appendix~\ref{sec:app:details:rl}).

The trajectories are also re-scaled in time such that all of them have the same duration and the action starts at a fixed timestamp.
We estimate the \emph{action phase}, the temporal part of the video where the actual action is performed by analyzing if the object is in motion (velocity $\dot{\bm o}_p$ above threshold) or if the hand is touching the object (predicted from~\cite{shan2020handobject}).
An \emph{approach phase} inserted at the beginning gives the gripper time to move to its initial position to execute the action.
A \emph{leave phase} at the end requires the gripper to be opened if the object should not remain grasped and the final position is set to be above the object position from the trajectory.


\subsection{Simulator and learning setup}
The simulated environment models the target robot environment that consists of a parallel jaw robotic gripper and cylindrical objects.
The gripper pose is controlled by linear and angular velocity and the gripper closing is specified by the gap between gripper fingers.
The simulation is performed for a fixed maximum horizon that corresponds to \SI{10}{\second} of simulated time and is prematurely reset if the gripper hits the ground.
When the environment resets, the poses of the hand and object(s) could be set according to the first state of the trajectory.
However, a policy learned with this constant starting state generalizes poorly to other starting conditions.
Therefore, we randomize the gripper starting poses and the object sizes at the beginning of each episode.
In particular, we employ automatic domain randomization, explained later in Sec.~\ref{subsec:eval:ablation}.

The goal of RL is to find a feed-forward policy that maps the observation to the distribution of actions that control the gripper.
We represent the policy~$\pi$ as a Gaussian distribution: $\pi(\bm a | \bm s) = \mathcal{N}(\bm \mu(\bm s), \Sigma)$,
where $\bm a \in \mathbb{R}^7$ is the gripper command,
$\bm s \in \mathbb{R}^{8}$ is the state containing time (1D) and gripper information (6D pose, 1D gripper opening),
$\bm \mu(\bm s) \in \mathbb{R}^7$ is a mapping represented by a neural network (three hidden linear layers of size 128),
and the diagonal covariance matrix $\Sigma \in \mathbb{R}^{7\times 7}$ is independent of the states.
The $\Sigma$ parameters and the neural network are trained by Proximal Policy Optimization~\cite{schulman2017ppo} to maximize the expected discounted return defined as
\begin{align}
    \label{eq:discounted_return}
    R = \mathbbm{E} \left[ \sum\nolimits_{i = 0}^H \gamma^i r(t_i) \right] \, ,
\end{align}
where~$t_i$ is timestamp at step~$i$,
horizon~$H$ specifies the maximum number of time steps,
$r$~is the immediate reward that encodes the distance to trajectories extracted from the videos (presented next),
and $\gamma$ is the discount factor, set to $1$ for our task to assign equal importance to future rewards.

\subsection{Learning a policy from single or multiple videos}
We start by defining the immediate reward for a single video as it forms the basis of multi-video learning.
Note that such policies are often brittle as small errors in state estimation can lead to irrecoverable errors in the policy.
For example, for the \emph{push} action, we expect the hand to be behind the object.
However, if this is not respected in the estimated trajectory, RL cannot recover from this error.
For learning from a single video, the immediate reward~$r_v$ at time $t$ is defined as:
\begin{align}
\label{eq:immediate_reward}
r_v(t) = \quad\sum_{\mathclap{\bm q \in \{ \bm h_p, \bm h_\theta, \bm o^i_p, \bm o^i_\theta, \tau \} }}\quad
w_q \exp \left( -\frac{1}{2} l_q d_q(\bm q(t), \tilde{\bm q}(t) )  \right) 
\, ,
\end{align}
where $\bm q$ is one of the quantities (hand position, object orientation, etc.) extracted from the video,
$\tilde{\bm q}$~is the corresponding quantity computed from the simulated environment,
function $d(\cdot, \cdot) $ measures distance between those quantities,
$l_q$ are constant length-scales used to compensate for differences in units, and
$w_q$ are constant weights used to adjust the importance of different quantities.
We assign the highest weight to object position as it is a crucial factor in judging the success of an action, and is difficult to learn as the policy may first need to learn other skills such as gripping the object.

We use different distance functions $d_q$ to compare simulated and estimated trajectories:
(i) squared euclidean norm for hand and object positions;
(ii) squared angular distance for hand azimuth and elevation;
(iii) squared angular distance between quaternions for object orientations; and
(iv) the following weak signal to learn gripper closing:
\begin{align}
    d(\tau, \tilde{\tau}) = 
    \begin{cases}
        \norm{\tau - \tilde{\tau}}^2  & \text{if reference touch signal detected in the video, \ie~$\tau = 1$,}\\
        \infty &\text{otherwise} \, .
    \end{cases}
\end{align}
This ensures that a positive reward is provided for closing the gripper only when the hand is touching the object, and the robot may arbitrarily choose to open/close the gripper at other timesteps.


\textbf{Reward for multiple videos.}
To learn a robust policy unaffected by errors in trajectory estimation, we propose to leverage multiple videos of the same action.
We train on $\mathcal{S}_a$, the subset of demonstrations corresponding to an action $a$, by averaging aforementioned immediate reward for a single video~\eqref{eq:immediate_reward}:
\begin{align}
\label{eq:return_multivideo}
r_a(t) = \frac{1}{|\mathcal{S}_a|} \sum\nolimits_{v \in \mathcal{S}_a } r_v(t) \, .
\end{align}
Note, that the average is computed over exponential rewards making it robust to outliers.
Even when there are multiple ways to achieve the same task, our policy learns to mimic the most common trajectory and not the averaged trajectory.
Rewards $r_v$~(Eq.~\ref{eq:immediate_reward}) and $r_a$~(Eq.~\ref{eq:return_multivideo}) are used as the immediate reward in Eq.~\eqref{eq:discounted_return} that is maximized by RL.

\section{Evaluation}
\label{sec:evaluation}

We first present an overview on the set of actions, a simulator-based benchmark, and metrics to evaluate the performance of learned policies.
We then present an ablation study highlighting the impact of visual recognition tools and a curriculum learning approach.
Finally, we compare policies learned on single or multiple videos and conclude with a brief summary of the real world robot setup.


\subsection{Evaluation setup}

\begin{figure}[t]
\centering
\includegraphics[width=\linewidth]{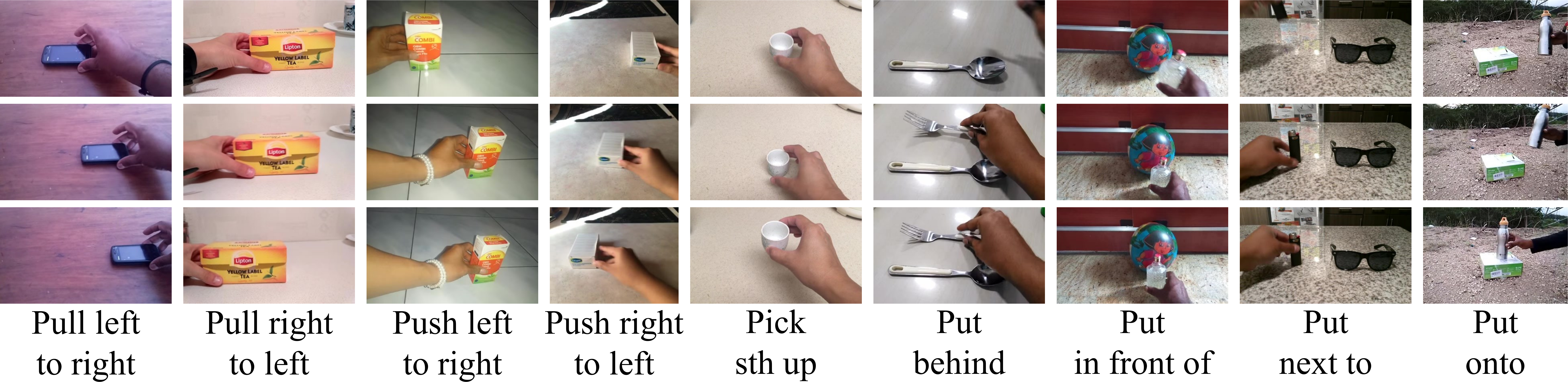}
\caption{
Example video demonstrations for the 9 actions used to evaluate our approach.
Three frames for each video depict the beginning, middle, and the end of the annotated \emph{action phase}.
}
\label{fig:actions}
\vspace{-5mm}
\end{figure}

\textbf{Actions.}
We evaluate our approach on nine actions from the Something-Something action recognition dataset~\cite{goyal2017something}:
five single object manipulation tasks (\emph{push/pull/pick})
and four two-object manipulation tasks (\emph{put}).
Fig.~\ref{fig:actions} illustrates one video per action category.
For each action, six diverse videos in terms of backgrounds and manipulated object categories and shapes are selected.

\textbf{Benchmark and metrics.}
We evaluate the robustness of trained policies by creating a benchmark of 1,000 samples that includes a variety of starting poses for the hand/gripper and sizes for the object(s).
This includes challenging setups that are not similar to video demonstrations, \eg~the hand starts behind the object.
We also create an easier benchmark with 1,000 samples drawn from a limited set of starting poses.
All results are presented on the \emph{hard} benchmark unless stated otherwise.

An estimated reward function is a proxy for the metric and is not always correlated with the successful execution of the action.
Therefore, we use hand designed action-specific metrics (only during evaluation) that analyze the state of the hand and object while executing the action.
Our stringent metric indicates a binary success/failure for each execution,
\eg~\emph{pull/push} actions check that the hand is in the correct position during object motion, that the object is standing, and it has moved at least \SI{50}{\milli\metre} in the correct direction.
Additional details are in Appendix~\ref{sec:app:benchmarks}.

\subsection{Experiments}
\label{subsec:eval:ablation}

\textbf{Real2Sim setups.}
We analyze the impact of failures in visual parsing and obtain 3D state trajectories in three ways:
A. ground-truth (GT) object/hand boxes~\cite{materzynska2020sthelse} and GT action phase localization;
B. GT boxes with predicted action phase; and
C. predicted boxes with predicted action phase.
Note that all setups still use predicted segmentation masks and automatic tracking.
As the videos are crowd-sourced for action recognition, there is no GT for the 3D states.
Thus, we use the action-specific metrics defined above to also compare the different Real2Sim setups.
The average success rate over all 54 videos for the three setups above is:
A. 67\%; B. 72\%; and C. 61\%.
Surprisingly, method B works best, possibly due to the trajectory being estimated for the complete video followed by action phase localization. 
Nevertheless, as we will see later, policies learn actions best with setup A.
Additional qualitative and quantitative analysis can be found in Appendix~\ref{sec:app:exp-real2sim}.

\begin{figure}[h]
\centering
\vspace{-2mm}
\begin{overpic}[width=\linewidth]{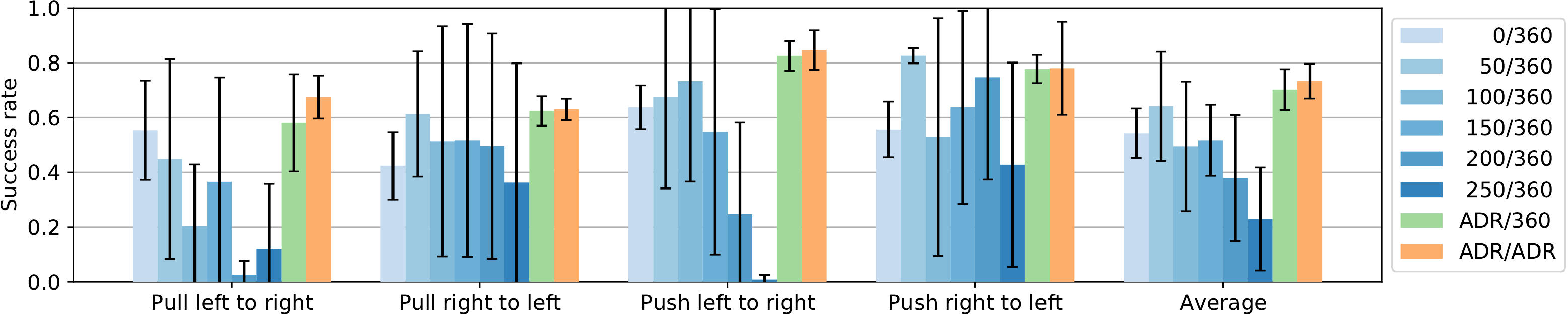}
 \put (90,1) {\scriptsize{position/angle}}
\end{overpic}
\caption{
Success rate for multi-video policies trained with different randomization strategies using state trajectories estimated from GT boxes/GT action phase.
\textbf{Legend}: the first number represents variability in the 3D position of the gripper, $\sigma$ in mm;
the second corresponds to the range of hand orientation in degrees.
Five policies are trained for each parameter and we report mean and standard deviation (error bar in the plot).
}
\vspace{-2mm}
\label{fig:curriculum}
\end{figure}

\textbf{Curriculum learning.}
We wish to learn policies that are robust to the initial gripper pose.
This is normally achieved by randomizing the starting gripper position and orientation during training, \ie,
$\bm h_p \sim \mathcal{N}(\bm 0, \sigma^2) \,$, $\bm h_\theta \sim \mathcal{U}(\ang{-180}, \ang{180})$,
where $\mathcal{N}$ and $\mathcal{U}$ are Normal and Uniform distributions, and $\sigma$ is set to \SI{250}{\milli\metre}.
However, the blue bars in Fig.~\ref{fig:curriculum} show that the performance drops as $\sigma$ increases indicating that learning with a randomized gripper pose is challenging.
Inspired by Automatic Domain Randomization (ADR)~\cite{akkaya2019solving}, we design a curriculum learning strategy that linearly increases the randomness of the gripper pose during training. 
Policies that use ADR for both the 3D position and orientation (Fig.~\ref{fig:curriculum} orange) achieve the best results and are used in all other experiments.

\begin{figure}[h]
\centering
\includegraphics[width=1.0\linewidth]{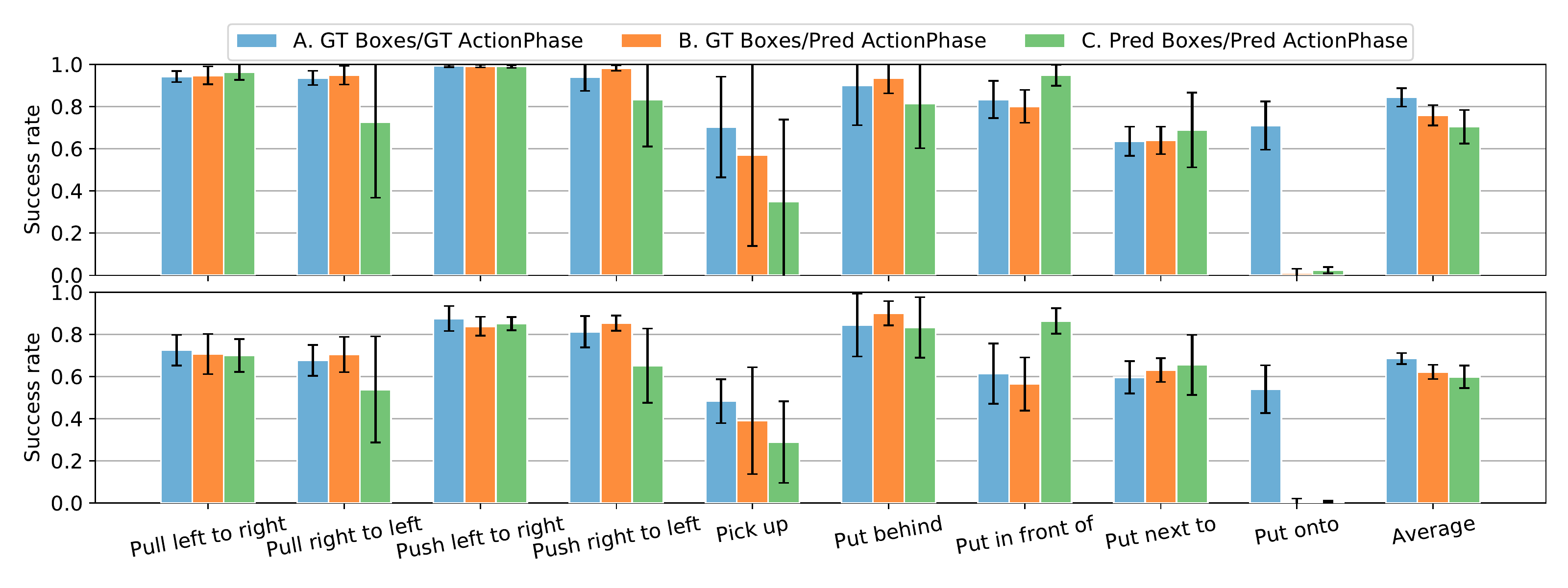}
\vspace{-6mm}
\caption{
Success rate for policies trained on multiple videos with trajectories estimated from different Real2Sim setups.
Top: easy benchmark; Bottom: hard benchmark.
Each bar represents mean and std-dev over ten policies.
}
\vspace{-3mm}
\label{fig:ablation}
\end{figure}

\textbf{Evaluating policies trained on different Real2Sim setups.}
Fig.~\ref{fig:ablation} shows the success rate for each action for policies trained on multiple videos evaluated on the easy (top) and hard (bottom) benchmarks.
While states predicted using method A (GT boxes and action phase, blue bars) result in higher performance, states that use method C (predicted boxes and action phase, green bars) degrade in performance by 10-15\% on both benchmarks.
Although method C results in higher performance for some actions (\eg~\emph{put in front of}), on average, the estimated vision works slightly worse than ground-truth as expected.
However, note that the action \emph{put onto} is found to be particularly challenging as it requires precise modeling of objects, and about 5-7\% of the gap between methods A and C is explained by this action alone.
Overall, the trained policies are successful at executing the actions for a majority of initial states on both benchmarks.
The gap between easy and hard benchmarks is also about 10-15\%, indicating the challenges in starting position and strictness of our metric.
A common failure case for our policy is when the gripper and object start close to each other and the gripper collides with the object while re-positioning in the approach phase.

\textbf{Learning from single vs. multiple videos.}
We demonstrate the benefits of learning a policy from multiple videos by comparing them against policies trained on individual videos for all actions.
In addition, we compare the proposed multi-video reward function to a baseline~\cite{peng2018deepMimic} that computes the reward by maximizing over the trajectories (instead of sum, see Eq.~\eqref{eq:return_multivideo}).
Table~\ref{table:single_vs_multi_sets5hard} shows that learning from single videos leads to a high variance in success rate due to the quality of the demonstration and state estimation process.
In contrast, our proposed approach learns the action even if only a few trajectories are correct (see \emph{put behind/in front of}).
Finally, the multi-video policy used in the baseline (DeepMimic~\cite{peng2018deepMimic}) results in lower performance than our approach on most actions.
The baseline computes maximum across trajectories, essentially selecting the easiest trajectory that would maximize reward. However, if this chosen trajectory does not successfully execute the action, it leads to negligible success rates.
as is seen for the relatively easier actions of \emph{pull left to right/right to left}.

\begin{table}[h]
        \centering
        \small
        \tabcolsep=1.2mm
        \begin{tabular}{l||cccccc||ccc}
        \toprule
Action &  vid 1 & vid 2 & vid 3 & vid 4 & vid 5 & vid 6 & average & proposed & baseline~\cite{peng2018deepMimic} \\ 
\midrule
Pull left to right & \cellcolor{gray!0} 83 & \cellcolor{gray!18} 68 & \cellcolor{gray!18} 67 & \cellcolor{gray!51} 46 & \cellcolor{gray!51} 45 & \cellcolor{gray!116} 2 & \cellcolor{gray!41} 52 & \cellcolor{gray!20} \textbf{66} & \cellcolor{gray!117} 1 \\
Pull right to left & \cellcolor{gray!26} 62 & \cellcolor{gray!31} 58 & \cellcolor{gray!42} 52 & \cellcolor{gray!99} 13 & \cellcolor{gray!120} 0 & \cellcolor{gray!120} 0 & \cellcolor{gray!73} 31 & \cellcolor{gray!60} \textbf{39} & \cellcolor{gray!119} 0 \\
Push left to right & \cellcolor{gray!0} 88 & \cellcolor{gray!0} 83 & \cellcolor{gray!18} 67 & \cellcolor{gray!33} 57 & \cellcolor{gray!57} 41 & \cellcolor{gray!120} 0 & \cellcolor{gray!35} 56 & \cellcolor{gray!0} \textbf{85} & \cellcolor{gray!119} 0 \\
Push right to left & \cellcolor{gray!0} 85 & \cellcolor{gray!0} 85 & \cellcolor{gray!9} 73 & \cellcolor{gray!13} 71 & \cellcolor{gray!28} 60 & \cellcolor{gray!31} 58 & \cellcolor{gray!11} 72 & \cellcolor{gray!9} \textbf{73} & \cellcolor{gray!13} 71 \\
Pick up & \cellcolor{gray!44} 50 & \cellcolor{gray!71} 32 & \cellcolor{gray!75} 29 & \cellcolor{gray!98} 14 & \cellcolor{gray!110} 6 & \cellcolor{gray!112} 5 & \cellcolor{gray!85} 22 & \cellcolor{gray!61} \textbf{38} & \cellcolor{gray!114} 3 \\
Put behind & \cellcolor{gray!0} 86 & \cellcolor{gray!0} 85 & \cellcolor{gray!34} 57 & \cellcolor{gray!51} 45 & \cellcolor{gray!66} 35 & \cellcolor{gray!75} 29 & \cellcolor{gray!35} 56 & \cellcolor{gray!0} \textbf{88} & \cellcolor{gray!0} 82 \\
Put in front of & \cellcolor{gray!0} 94 & \cellcolor{gray!0} 80 & \cellcolor{gray!39} 54 & \cellcolor{gray!49} 46 & \cellcolor{gray!66} 35 & \cellcolor{gray!92} 18 & \cellcolor{gray!37} 54 & \cellcolor{gray!0} \textbf{83} & \cellcolor{gray!22} 65 \\
Put next to & \cellcolor{gray!13} 71 & \cellcolor{gray!16} 68 & \cellcolor{gray!31} 58 & \cellcolor{gray!34} 56 & \cellcolor{gray!39} 53 & \cellcolor{gray!44} 50 & \cellcolor{gray!30} 59 & \cellcolor{gray!37} 55 & \cellcolor{gray!2} \textbf{78} \\
Put onto & \cellcolor{gray!91} 19 & \cellcolor{gray!119} 0 & \cellcolor{gray!120} 0 & \cellcolor{gray!118} 0 & \cellcolor{gray!119} 0 & \cellcolor{gray!120} 0 & \cellcolor{gray!114} \textbf{3} & \cellcolor{gray!119} 0 & \cellcolor{gray!119} 0 \\
\bottomrule 
\end{tabular}
\vspace{1mm}
\caption{
The success rate (in \%) for single and multi-video policies and for the baseline~\cite{peng2018deepMimic}.
Single video performances are sorted in descending order, and their average score is presented in the ``average'' column.
Results are on the hard benchmark with states from method C, predicted boxes and action phase.
Please refer to Appendix~\ref{sec:app:exp-policy:single-vs-multi} for similar results on the \emph{easy} benchmark, and with states estimated from method A.
}
\vspace{-4mm}
\label{table:single_vs_multi_sets5hard} 
\end{table}

\textbf{Transferring learned skills to the real robot.}
As the trained policy predicts the linear and angular velocity and the amount of gripper opening based on the current state of the gripper, we can compute the whole execution trajectory offline for an object with known initial pose (required for the alignment, and assumed known for our setup).
We use numerical inverse kinematics tracking to compute the robot joint trajectory from the gripper Cartesian trajectory and quantize the gripper opening/closing due to lack of real-time gripper control capability for our Franka Emika Panda robot, shown in Fig.~\ref{fig:teaser}.
Please refer to the supplementary video for a demo.

\section{Conclusion}\label{sec:conclusion}

We presented an approach to teach robotic agents simple object manipulation skills by watching a few videos.
We proposed a method that estimates a coarse 3D state representation for the hand and object(s) through a combination of 2D visual recognition, differentiable rendering, and an optimization method that learns from perceptual and physics based losses.
These approximate state trajectories are used in an RL setup to successfully learn object manipulations for 9 single- and multi-object actions.
We performed a thorough evaluation in a simulated environment, highlighting the benefits of adopting a curriculum learning strategy and learning from multiple videos, and also showed that the learned policies can be transferred to a real robot.
Interesting future directions include incorporating physics into the Real2Sim estimation and scaling up to more actions.

\textbf{Acknowledgment.}
This work was funded in part by the Louis Vuitton ENS Chair on Artificial Intelligence, the French government under management of Agence Nationale de la Recherche as part of the ``Investissements d'avenir'' program, reference ANR-19-P3IA-0001 (PRAIRIE 3IA Institute), and the European Regional Development Fund under the project IMPACT (reg. no. CZ.02.1.01/0.0/0.0/15\_003/0000468).

{
\small{}
\clearpage
\bibliography{shortstrings,refs}

\begin{thebibliography}{45}
\providecommand{\natexlab}[1]{#1}
\providecommand{\url}[1]{\texttt{#1}}
\expandafter\ifx\csname urlstyle\endcsname\relax
  \providecommand{\doi}[1]{doi: #1}\else
  \providecommand{\doi}{doi: \begingroup \urlstyle{rm}\Url}\fi

\bibitem[Miech et~al.(2019)Miech, Zhukov, Alayrac, Tapaswi, Laptev, and
  Sivic]{miech2019howto100m}
A.~Miech, D.~Zhukov, J.-B. Alayrac, M.~Tapaswi, I.~Laptev, and J.~Sivic.
\newblock {HowTo100M: Learning a Text-Video Embedding by Watching Hundred
  Million Narrated Video Clips}.
\newblock In \emph{ICCV}, 2019.

\bibitem[Argall et~al.(2009)Argall, Chernova, Veloso, and
  Browning]{argall2009survey}
B.~D. Argall, S.~Chernova, M.~Veloso, and B.~Browning.
\newblock {A survey of robot learning from demonstration}.
\newblock \emph{Robotics and Autonomous Systems}, 57\penalty0 (5):\penalty0
  469--483, 2009.

\bibitem[Zhang et~al.(2018)Zhang, McCarthy, Jow, Lee, Chen, Goldberg, and
  Abbeel]{zhang2018imitationComplexTasks}
T.~Zhang, Z.~McCarthy, O.~Jow, D.~Lee, X.~Chen, K.~Goldberg, and P.~Abbeel.
\newblock Deep imitation learning for complex manipulation tasks from virtual
  reality teleoperation.
\newblock In \emph{ICRA}, 2018.

\bibitem[Hasson et~al.(2019)Hasson, Varol, Tzionas, Kalevatykh, Black, Laptev,
  and Schmid]{hasson2019obman}
Y.~Hasson, G.~Varol, D.~Tzionas, I.~Kalevatykh, M.~J. Black, I.~Laptev, and
  C.~Schmid.
\newblock Learning joint reconstruction of hands and manipulated objects.
\newblock In \emph{CVPR}, 2019.

\bibitem[Shan et~al.(2020)Shan, Geng, Shu, and Fouhey]{shan2020handobject}
D.~Shan, J.~Geng, M.~Shu, and D.~Fouhey.
\newblock {Understanding Human Hands in Contact at Internet Scale}.
\newblock In \emph{CVPR}, 2020.

\bibitem[He et~al.(2017)He, Gkioxari, Dollar, and Girshick]{he2017maskrcnn}
K.~He, G.~Gkioxari, P.~Dollar, and R.~Girshick.
\newblock {Mask R-CNN}.
\newblock In \emph{ICCV}, 2017.

\bibitem[Goyal et~al.(2017)Goyal, Kahou, Michalski, Materzyńska, Westphal,
  Kim, Haenel, Fruend, Yianilos, Mueller-Freitag, Hoppe, Thurau, Bax, and
  Memisevic]{goyal2017something}
R.~Goyal, S.~E. Kahou, V.~Michalski, J.~Materzyńska, S.~Westphal, H.~Kim,
  V.~Haenel, I.~Fruend, P.~Yianilos, M.~Mueller-Freitag, F.~Hoppe, C.~Thurau,
  I.~Bax, and R.~Memisevic.
\newblock {The ``something something'' video database for learning and
  evaluating visual common sense}.
\newblock In \emph{ICCV}, 2017.

\bibitem[Akkaya et~al.(2019)Akkaya, Andrychowicz, Chociej, Litwin, McGrew,
  Petron, Paino, Plappert, Powell, Ribas, et~al.]{akkaya2019solving}
I.~Akkaya, M.~Andrychowicz, M.~Chociej, M.~Litwin, B.~McGrew, A.~Petron,
  A.~Paino, M.~Plappert, G.~Powell, R.~Ribas, et~al.
\newblock Solving rubik's cube with a robot hand.
\newblock \emph{arXiv preprint arXiv:1910.07113}, 2019.

\bibitem[Roberts(1963)]{roberts1963blocksworld}
L.~G. Roberts.
\newblock \emph{{Machine perception of three-dimensional solids}}.
\newblock PhD thesis, Massachusetts Institute of Technology, 1963.

\bibitem[Gupta et~al.(2010)Gupta, Efros, and Hebert]{gupta2010blocksworld}
A.~Gupta, A.~A. Efros, and M.~Hebert.
\newblock {Blocks World Revisited: Image Understanding Using Qualitative
  Geometry and Mechanics}.
\newblock In \emph{ECCV}, 2010.

\bibitem[Gkioxari et~al.(2019)Gkioxari, Malik, and
  Johnson]{gkioxari2019meshrcnn}
G.~Gkioxari, J.~Malik, and J.~Johnson.
\newblock {Mesh R-CNN}.
\newblock In \emph{ICCV}, 2019.

\bibitem[Wu et~al.(2017{\natexlab{a}})Wu, Tenenbaum, and Kohli]{wu2017nsd}
J.~Wu, J.~B. Tenenbaum, and P.~Kohli.
\newblock Neural scene de-rendering.
\newblock In \emph{CVPR}, 2017{\natexlab{a}}.

\bibitem[Wu et~al.(2017{\natexlab{b}})Wu, Lu, Kohli, Freeman, and
  Tenenbaum]{wu2017vsd}
J.~Wu, E.~Lu, P.~Kohli, B.~Freeman, and J.~Tenenbaum.
\newblock Learning to see physics via visual de-animation.
\newblock In \emph{NeurIPS}, 2017{\natexlab{b}}.

\bibitem[Wang et~al.(2018)Wang, Zhang, Li, Fu, Liu, and
  Jiang]{wang2018pixel2mesh}
N.~Wang, Y.~Zhang, Z.~Li, Y.~Fu, W.~Liu, and Y.-G. Jiang.
\newblock {Pixel2Mesh: Generating 3D Mesh Models from Single RGB Images}.
\newblock In \emph{ECCV}, 2018.

\bibitem[Kato et~al.(2018)Kato, Ushiku, and Harada]{kato2018renderer}
H.~Kato, Y.~Ushiku, and T.~Harada.
\newblock {Neural 3D Mesh Renderer}.
\newblock In \emph{CVPR}, 2018.

\bibitem[Liu et~al.(2019)Liu, Li, Chen, and Li]{liu2019softrasterizer}
S.~Liu, T.~Li, W.~Chen, and H.~Li.
\newblock {Soft Rasterizer: A Differentiable Renderer for Image-based 3D
  Reasoning}.
\newblock In \emph{ICCV}, 2019.

\bibitem[Chen et~al.(2019)Chen, Gao, Ling, Smith, Lehtinen, Jacobson, and
  Fidler]{chen2019dib}
W.~Chen, J.~Gao, H.~Ling, E.~J. Smith, J.~Lehtinen, A.~Jacobson, and S.~Fidler.
\newblock {Learning to Predict 3D Objects with an Interpolation-based
  Differentiable Renderer}.
\newblock In \emph{NeurIPS}, 2019.

\bibitem[Damen et~al.(2020)Damen, Doughty, Farinella, Fidler, Furnari, Kazakos,
  Moltisanti, Munro, Perrett, Price, and Wray]{damen2020epickitchens}
D.~Damen, H.~Doughty, G.~M. Farinella, S.~Fidler, A.~Furnari, E.~Kazakos,
  D.~Moltisanti, J.~Munro, T.~Perrett, W.~Price, and M.~Wray.
\newblock {The EPIC-KITCHENS Dataset: Collection, Challenges and Baselines}.
\newblock \emph{PAMI}, 2020.

\bibitem[Tekin et~al.(2019)Tekin, Bogo, and Pollefeys]{tekin2019hoact}
B.~Tekin, F.~Bogo, and M.~Pollefeys.
\newblock {H+O: Unified Egocentric Recognition of 3D Hand-Object Poses and
  Interactions}.
\newblock In \emph{CVPR}, 2019.

\bibitem[Höll et~al.(2018)Höll, Oberweger, Arth, and Lepetit]{holl2018vrhand}
M.~Höll, M.~Oberweger, C.~Arth, and V.~Lepetit.
\newblock {Efficient Physics-Based Implementation for Realistic Hand-Object
  Interaction in Virtual Reality}.
\newblock In \emph{IEEE Conference on Virtual Reality and 3D User Interfaces
  (VR)}, 2018.

\bibitem[Tang et~al.(2020)Tang, Hu, Fu, and Cohen-Or]{tang2020arhands}
X.~Tang, X.~Hu, C.-W. Fu, and D.~Cohen-Or.
\newblock {GrabAR: Occlusion-aware Grabbing Virtual Objects in AR}.
\newblock In \emph{ACM User Interface Software and Technology Symposium}, 2020.

\bibitem[Bogo et~al.(2016)Bogo, Kanazawa, Lassner, Gehler, Romero, and
  Black]{bogo2016smpl}
F.~Bogo, A.~Kanazawa, C.~Lassner, P.~Gehler, J.~Romero, and M.~J. Black.
\newblock {Keep it SMPL: Automatic Estimation of 3D Human Pose and Shape from a
  Single Image}.
\newblock In \emph{ECCV}, 2016.

\bibitem[Romero et~al.(2017)Romero, Tzionas, and Black]{romero2017mano}
J.~Romero, D.~Tzionas, and M.~J. Black.
\newblock {Embodied Hands: Modeling and Capturing Hands and Bodies Together}.
\newblock \emph{ACM Transactions on Graphics, (Proc. SIGGRAPH Asia)},
  36\penalty0 (6):\penalty0 245:1--245:17, 2017.

\bibitem[Zhou et~al.(2020)Zhou, Habermann, Xu, Habibie, Theobalt, and
  Xu]{zhou2020monocular}
Y.~Zhou, M.~Habermann, W.~Xu, I.~Habibie, C.~Theobalt, and F.~Xu.
\newblock {Monocular Real-time Hand Shape and Motion Capture using Multi-modal
  Data}.
\newblock In \emph{CVPR}, 2020.

\bibitem[Materzynska et~al.(2020)Materzynska, Xiao, Herzig, and
  Xu]{materzynska2020sthelse}
J.~Materzynska, T.~Xiao, R.~Herzig, and H.~Xu.
\newblock {Something-Else: Compositional Action Recognition with
  Spatial-Temporal Interaction Networks}.
\newblock In \emph{CVPR}, 2020.

\bibitem[Hampali et~al.(2020)Hampali, Rad, Oberweger, and
  Lepetit]{hampali2020honnotate}
S.~Hampali, M.~Rad, M.~Oberweger, and V.~Lepetit.
\newblock {HOnnotate: A method for 3D Annotation of Hand and Object Poses}.
\newblock In \emph{CVPR}, 2020.

\bibitem[Hasson et~al.(2020)Hasson, Tekin, Bogo, Laptev, Pollefeys, and
  Schmid]{hasson2020handobjecttime}
Y.~Hasson, B.~Tekin, F.~Bogo, I.~Laptev, M.~Pollefeys, and C.~Schmid.
\newblock {Leveraging Photometric Consistency over Time for Sparsely Supervised
  Hand-Object Reconstruction}.
\newblock In \emph{CVPR}, 2020.

\bibitem[Hazara and Kyrki(2016)]{hazara2016contact}
M.~Hazara and V.~Kyrki.
\newblock {Reinforcement learning for improving imitated in-contact skills}.
\newblock In \emph{International Conference on Humanoid Robots}, 2016.

\bibitem[Vecerik et~al.(2017)Vecerik, Hester, Scholz, Wang, Pietquin, Piot,
  Heess, Rothörl, Lampe, and Riedmiller]{vecerik2017demonstrations}
M.~Vecerik, T.~Hester, J.~Scholz, F.~Wang, O.~Pietquin, B.~Piot, N.~Heess,
  T.~Rothörl, T.~Lampe, and M.~Riedmiller.
\newblock {Leveraging Demonstrations for Deep Reinforcement Learning on
  Robotics Problems with Sparse Rewards}.
\newblock \emph{arXiv: 1707.08817}, 2017.

\bibitem[Yu et~al.(2019)Yu, Quillen, He, Julian, Hausman, Finn, and
  Levine]{yu2019metaworld}
T.~Yu, D.~Quillen, Z.~He, R.~Julian, K.~Hausman, C.~Finn, and S.~Levine.
\newblock {Meta-World: A Benchmark and Evaluation for Multi-Task and Meta
  Reinforcement Learning}.
\newblock In \emph{CoRL}, 2019.

\bibitem[Caccavale et~al.(2019)Caccavale, Saveriano, Finzi, and
  Lee]{caccavale2019kinesthetic}
R.~Caccavale, M.~Saveriano, A.~Finzi, and D.~Lee.
\newblock {Kinesthetic teaching and attentional supervision of structured tasks
  in human-robot interaction}.
\newblock \emph{Autonomous Robots}, 43:\penalty0 1291--1307, 2019.

\bibitem[Strudel et~al.(2020)Strudel, Pashevich, Kalevatykh, Laptev, Sivic, and
  Schmid]{strudel2020rlbc}
R.~Strudel, A.~Pashevich, I.~Kalevatykh, I.~Laptev, J.~Sivic, and C.~Schmid.
\newblock Learning to combine primitive skills: A step towards versatile
  robotic manipulation.
\newblock In \emph{ICRA}, 2020.

\bibitem[Sermanet et~al.(2017{\natexlab{a}})Sermanet, Xu, and
  Levine]{sermanet2017perceptualrewards}
P.~Sermanet, K.~Xu, and S.~Levine.
\newblock {Unsupervised Perceptual Rewards for Imitation Learning}.
\newblock In \emph{RSS}, 2017{\natexlab{a}}.

\bibitem[Sermanet et~al.(2017{\natexlab{b}})Sermanet, Lynch, Chebotar, Hsu,
  Jang, Schaal, and Levine]{sermanet2017tcn}
P.~Sermanet, C.~Lynch, Y.~Chebotar, J.~Hsu, E.~Jang, S.~Schaal, and S.~Levine.
\newblock {Time-Contrastive Networks: Self-Supervised Learning from Video}.
\newblock In \emph{ICRA}, 2017{\natexlab{b}}.

\bibitem[Liu et~al.(2018)Liu, Gupta, Abbeel, and
  Levine]{liu2018imitationObservation}
Y.~Liu, A.~Gupta, P.~Abbeel, and S.~Levine.
\newblock {Imitation from Observation}: Learning to imitate behaviors from raw
  video via context translation.
\newblock In \emph{ICRA}, 2018.

\bibitem[Smith et~al.(2020)Smith, Dhawan, Zhang, Abbeel, and
  Levine]{smith2020avid}
L.~Smith, N.~Dhawan, M.~Zhang, P.~Abbeel, and S.~Levine.
\newblock {AVID: Learning Multi-Stage Tasks via Pixel-Level Translation of
  Human Videos}.
\newblock In \emph{RSS}, 2020.

\bibitem[Shao et~al.(2020)Shao, Migimatsu, Zhang, Yang, and
  Bohg]{shao2020concept2robot}
L.~Shao, T.~Migimatsu, Q.~Zhang, K.~Yang, and J.~Bohg.
\newblock {Concept2Robot: Learning Manipulation Concepts from Instructions and
  Human Demonstrations}.
\newblock In \emph{RSS}, 2020.

\bibitem[Kanazawa et~al.(2018)Kanazawa, Black, Jacobs, and
  Malik]{kanazawa2018hmr}
A.~Kanazawa, M.~J. Black, D.~W. Jacobs, and J.~Malik.
\newblock {End-to-end Recovery of Human Shape and Pose}.
\newblock In \emph{CVPR}, 2018.

\bibitem[Peng et~al.(2018{\natexlab{a}})Peng, Abbeel, Levine, and van~de
  Panne]{peng2018deepMimic}
X.~B. Peng, P.~Abbeel, S.~Levine, and M.~van~de Panne.
\newblock {DeepMimic: Example-guided Deep Reinforcement Learning of
  Physics-based Character Skills}.
\newblock \emph{ACM Trans. Graph.}, 37\penalty0 (4):\penalty0 143:1--143:14,
  2018{\natexlab{a}}.

\bibitem[Peng et~al.(2018{\natexlab{b}})Peng, Kanazawa, Malik, Abbeel, and
  Levine]{peng2018sfv}
X.~B. Peng, A.~Kanazawa, J.~Malik, P.~Abbeel, and S.~Levine.
\newblock {SFV: Reinforcement Learning of Physical Skills from Videos}.
\newblock \emph{ACM Trans. Graph.}, 37\penalty0 (6), 2018{\natexlab{b}}.

\bibitem[Peng et~al.(2020)Peng, Coumans, Zhang, Lee, Tan, and
  Levine]{peng2020imitatingAnimals}
X.~B. Peng, E.~Coumans, T.~Zhang, T.-W. Lee, J.~Tan, and S.~Levine.
\newblock Learning agile robotic locomotion skills by imitating animals.
\newblock In \emph{RSS}, 2020.

\bibitem[Dawson-Haggerty et~al.(2019)]{trimesh}
Dawson-Haggerty et~al.
\newblock trimesh, 2019.
\newblock URL \url{https://trimsh.org/}.

\bibitem[Schulman et~al.(2017)Schulman, Wolski, Dhariwal, Radford, and
  Klimov]{schulman2017ppo}
J.~Schulman, F.~Wolski, P.~Dhariwal, A.~Radford, and O.~Klimov.
\newblock {Proximal Policy Optimization Algorithms}.
\newblock \emph{arXiv:1707.06347}, 2017.

\bibitem[Lin et~al.(2014)Lin, Maire, Belongie, Hays, Perona, Ramanan, Dollár,
  and Zitnick]{lin2014coco}
T.-Y. Lin, M.~Maire, S.~Belongie, J.~Hays, P.~Perona, D.~Ramanan, P.~Dollár,
  and C.~L. Zitnick.
\newblock {Microsoft COCO: Common Objects in Context}.
\newblock In \emph{ECCV}, 2014.

\bibitem[Bewley et~al.(2016)Bewley, Ge, Ott, Ramos, and
  Upcroft]{bewley2016sort}
A.~Bewley, Z.~Ge, L.~Ott, F.~Ramos, and B.~Upcroft.
\newblock {Simple Online and Realtime Tracking}.
\newblock In \emph{ICIP}, 2016.

\end{thebibliography}
}

\clearpage
\appendix
\section*{Appendix}
We present additional details for our method and a variety of qualitative and quantitative results.
This appendix is complemented by additional material including a video describing the contributions in brief and demonstrating results on a real robot, and two HTML files: \texttt{real2sim.html} and \texttt{policies.html} showcasing qualitative real2sim and policy results respectively. 
\FloatBarrier\section{Implementation Details}

We present some additional implementation details for
(i) \emph{Video pre-processing} that estimates hand and object detections and segmentation masks;
(ii) \emph{Real2Sim}, that estimates 3D state trajectories from videos; and
(iii) \emph{RL} that learns a policy to mimic object manipulation based on extracted trajectories.

\subsection{Video pre-processing}
\label{sec:app:details:preproc}
We parse each video with a frame-level hand-object detector and semantic segmentation method.

(i) Mask-RCNN~\cite{he2017maskrcnn} predicts segmentation masks for multiple COCO categories~\cite{lin2014coco}, including the \emph{person} label that segments the hand, and various objects that may appear in the video.

(ii) Hand-Object detector~\cite{shan2020handobject} is a recent work that detects hands, objects, and their interactions (touching or not).
As an alternative to automatic detection, we also analyze the impact of using ground-truth boxes (Something-Else~\cite{materzynska2020sthelse}).
However, note that this still uses automatic segmentation to convert boxes to pixel-level masks.

\textbf{Tracking.}
We track the hand and object(s) of interest by adopting a Kalman-filter~\cite{bewley2016sort} that links frame-level detections using a simple intersection-over-union overlap based metric.
Missed detections are ignored, and the segmentation mask is set to all zero.
The physics-based losses, especially the acceleration loss, interpolates across missed detections by preventing rapid changes in position and orientation.

\subsection{Real2Sim}
\label{sec:app:details:real2sim}
\textbf{3D state estimation parameters.}
States are represented as learnable parameters (real numbers) that are normalized through sigmoid or tanh functions.

We initialize 3D state parameters of the hand (3D position, elevation) and the object (3D size and 3D position) by sampling from a normal distribution: $0.1 \cdot \mathcal{N}(0, 1)$.
Hand azimuth is initialized from $0.1 \cdot \mathcal{N}(\pi/2, 1)$ to model the hand orientation in egocentric videos (pointing away from the camera at \ang{90}), and finally
object rotation is initialized as $0.01 \cdot \mathcal{N}(0, 1)$ as we expect objects to be upright.

We normalize these parameters as follows:
(i) object sizes are normalized through a sigmoid function in range \SIrange{0}{300}{\milli\metre};
(ii) object position is normalized through a tanh function in range \SIrange{-1200}{1200}{\milli\metre};
(iii) hand position is normalized through a tanh function in range
\SIrange{-1500}{1500}{\milli\metre}; and
(iv) hand elevation is normalized using a tanh function in range \ang{-90} to \ang{90}.
We found that normalizing hand azimuth makes state estimation challenging, possibly due to \ang{0} and \ang{360} corresponding to the same angle.
The object position in z direction is set to half the object size in z direction as objects may lie on the table.

\textbf{Weights for the loss terms}
are as follows.
For the perceptual loss $w_p = 0.3$,
acceleration term $w_a = 5$,
interaction term $w_i = 1$, and
object size term $w_s = 1000$.
Note that the high weight $w_s$ ensures that the object size does not change over the estimated trajectory.

\textbf{Learning details.}
We use the Adam optimizer with a (considerably high) learning rate $10^{-2}$ to update the state parameters.
All states for the entire video are updated at each iteration as this allows imposing acceleration based regularization at every frame.
The parameters are updated for 400 iterations (most parameters converge typically around 200-250 iterations).
The models are implemented using PyTorch.

\textbf{Action phase estimation.}
Given the entire 3D state trajectory, we use two cues to predict when the action is taking place.
First, we consider whether the hand is touching the object~\cite{shan2020handobject}.
Second, we consider whether the object is in motion, by checking if the velocity is greater than \SI{70}{\milli\metre\per\second}.
These predictions are quite reliable with an average intersection-over-union of 0.81.

\subsection{Learning object manipulation policy from multiple videos}
\label{sec:app:details:rl}
\begin{figure}[h]
\centering
\begin{overpic}[height=2cm,abs,unit=1mm]{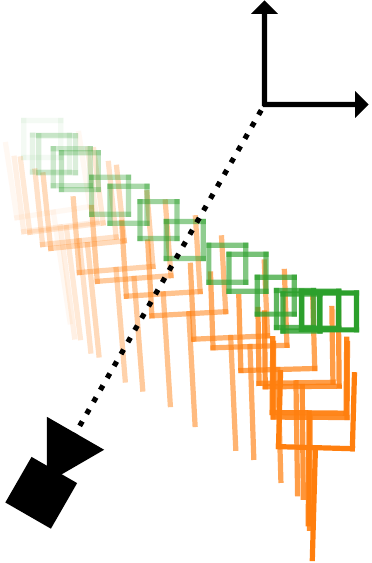}
 \put (-5,16) {(a)}
\end{overpic}
\hspace{1.5cm}
\begin{overpic}[height=2cm,abs,unit=1mm]{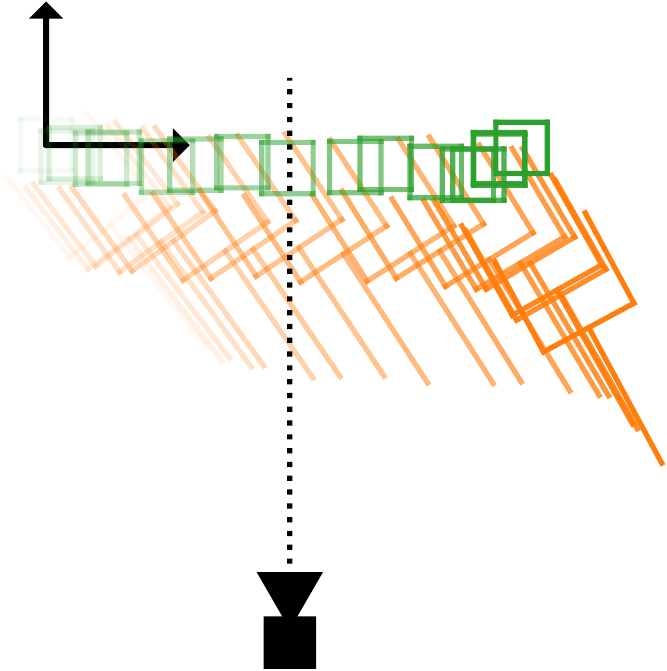}
 \put (-5,16) {(b)}
\end{overpic}
\hspace{1.5cm}
\begin{overpic}[height=2cm,abs,unit=1mm]{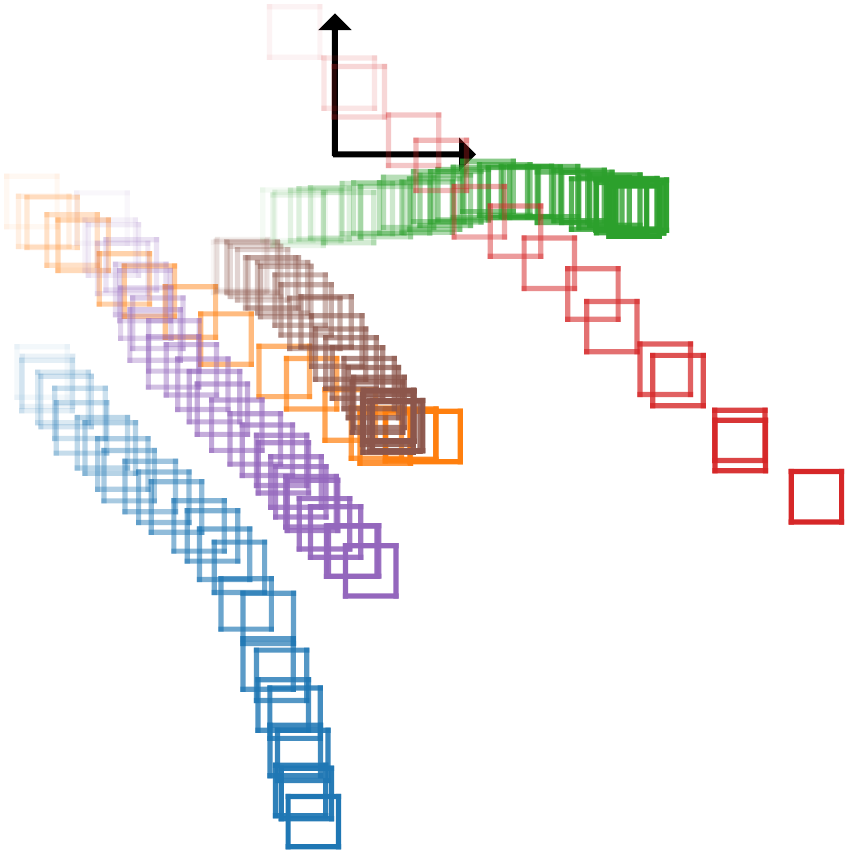}
 \put (-5,16) {(c)}
\end{overpic}
\hspace{1.5cm}
\begin{overpic}[height=2cm,abs,unit=1mm]{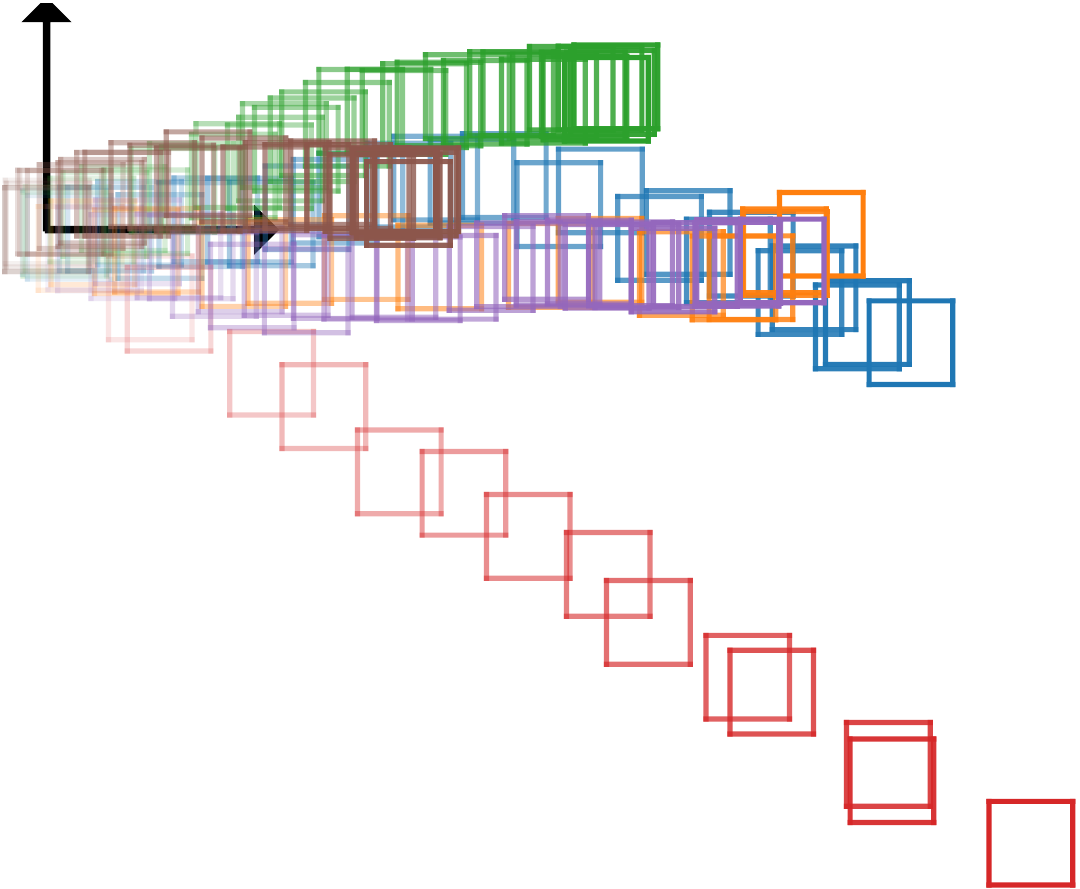}
 \put (-5,16) {(d)}
\end{overpic}
\caption{
Top-view explaining the spatial alignment of extracted trajectories for the action \emph{pull left to right}.
Time is indicated by transparency, the most transparent color represents the start of the trajectory.
For one video, \textbf{(a)} shows the estimated camera pose, hand (orange), and object (green) trajectories, while \textbf{(b)} shows the trajectories after alignment, compensating for camera pose and initial object position.
\textbf{(c)} depicts object trajectories extracted from multiple videos of the same action, and \textbf{(d)} presents their aligned version.
}
\vspace{-2mm}
\label{fig:alignment}
\end{figure}

\textbf{Spatial alignment.}
We spatially align all trajectories of the same action by compensating for the camera orientation and the initial object position as shown in Fig.~\ref{fig:alignment}.
The spatial alignment merely resets the frame-of-reference (camera and initial object position) to an object in the scene, something that can be achieved for most actions.

\textbf{Reinforcement learning parameters.}
Reward weights required for computation of Eq.~\eqref{eq:immediate_reward} are shown in Table~\ref{tab:rewards_rl}.
We use PPO~\cite{schulman2017ppo} algorithm that collects 70~episodes per policy update.
Policy update uses learning rate~$3\cdot 10^{-4}$ and is performed for 25~epochs, value 0.2~for likelihood ratio clipping threshold, value 100~for gradient clipping, weight~$1\cdot 10^{-8}$ for entropy loss and generalized advantage estimation with lambda set to~0.95.
Discount factor is set to 1.
Total number of policy updates is set to 1,000.

For Automatic Domain Randomization (ADR), we linearly increases randomness for gripper orientation from the 1st until the 400th policy update and for gripper position from 500th until the 900th policy update.
The policy is represented by a fully connected neural network with three hidden layers each containing 128~neurons.
The models are implemented using PyTorch and \emph{rlpyt}, a RL library for PyTorch.

\begin{table}[h]
    \centering
    \tabcolsep=1.2mm

    \begin{tabular}{cccccc}
        \toprule
        Parameter & $\bm h_p$ & $\bm h_\theta$ & $\bm o^i_p$ & $\bm o^i_\theta$ & $\tau$\\
        \midrule
        $w_q$ & 0.2 & 0.2 & 0.5 & 0.05 & 0.05 \\
        $l_q$ & 100 & 10 & 100 & 10 & 10000 \\
        \bottomrule
    \end{tabular}
    \caption{Reward parameters for RL.}
    \label{tab:rewards_rl}
\end{table}

\FloatBarrier\section{Benchmark and Metrics}
\label{sec:app:benchmarks}

The \textit{easy} and \textit{hard} benchmarks are sampled randomly from distributions specified in Table~\ref{tab:benchmark_ranges}.
The random sample is discarded if a collision is detected among any two of hand, object(s), or ground plane.
In total 1,000~collision free samples are obtained for each benchmark and used for the evaluation.

\begin{table}[h]
    \centering
    \begin{tabular}{lll}
        \toprule
        Parameter & \textit{easy} benchmark & \textit{hard} benchmark \\
        \midrule
        gripper position x & $\mathcal{U}(\SI{-100}{\milli\metre}, \SI{100}{\milli\metre})$ & $\mathcal{U}(\SI{-250}{\milli\metre}, \SI{250}{\milli\metre})$ \\
        gripper position y & $\mathcal{U}(\SI{-100}{\milli\metre}, \SI{0}{\milli\metre})$ & $\mathcal{U}(\SI{-250}{\milli\metre}, \SI{250}{\milli\metre})$ \\
        gripper position z & $\mathcal{U}(\SI{100}{\milli\metre}, \SI{200}{\milli\metre})$ & $\mathcal{U}(\SI{0}{\milli\metre}, \SI{250}{\milli\metre})$ \\
        gripper azimuth & $\mathcal{U}(\ang{-180}, \ang{180})$ & $\mathcal{U}(\ang{-180}, \ang{180})$ \\
        gripper elevation & $\mathcal{U}(\ang{0}, \ang{80})$ & $\mathcal{U}(\ang{-80}, \ang{80})$ \\
        1st object position & $\bm 0$ & $\bm 0$ \\
        1st object orientation & upright & upright \\
        1st object radius & $\mathcal{U}(\SI{40}{\milli\metre}, \SI{50}{\milli\metre})$ & $\mathcal{U}(\SI{40}{\milli\metre}, \SI{50}{\milli\metre})$ \\
        1st object height & $\mathcal{U}(\SI{60}{\milli\metre}, \SI{80}{\milli\metre})$ & $\mathcal{U}(\SI{40}{\milli\metre}, \SI{100}{\milli\metre})$ \\
        2nd object pose & in hand & in hand \\
        2nd object radius & $\mathcal{U}(\SI{40}{\milli\metre}, \SI{50}{\milli\metre})$ & $\mathcal{U}(\SI{40}{\milli\metre}, \SI{50}{\milli\metre})$ \\
        2nd object height & $\mathcal{U}(\SI{60}{\milli\metre}, \SI{80}{\milli\metre})$ & $\mathcal{U}(\SI{40}{\milli\metre}, \SI{100}{\milli\metre})$ \\
        \bottomrule
    \end{tabular}
    \caption{
        Ranges for benchmark generation.
        Symbol~$\mathcal{U}$ represents uniform distribution.
    }
    \label{tab:benchmark_ranges}
\end{table}

%

\begin{figure}[h]
\centering
\includegraphics[width=\linewidth]{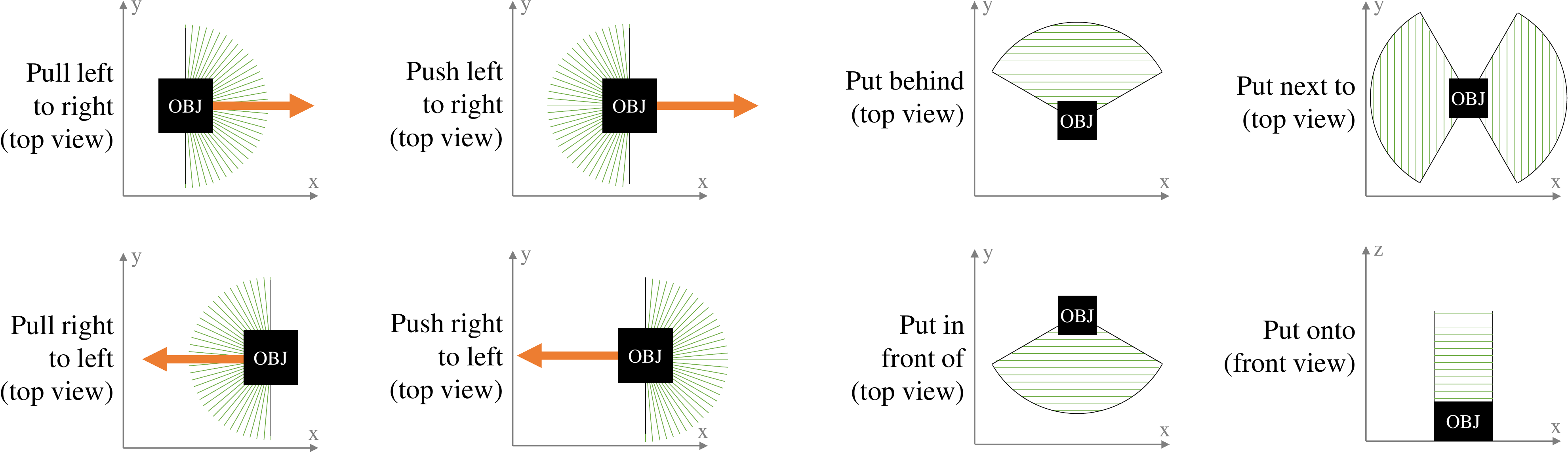}
\caption{Visualization of the metrics used to evaluate whether the action is performed correctly.
Note how the hand is in the direction of motion for pull actions, while in the opposite area for the push actions.
Two object actions allow the object to be \emph{put} in a specific cone of area \emph{behind/in front of/next to} the object.
For \emph{put onto}, we show the front view, indicating that the second object should be placed above the static one.
Please refer to the text for exact numerical details.
}
\label{fig:metrics}
\end{figure}

\textbf{Metrics.}
We design an evaluation metric for each action separately by analyzing the positions and orientations of the hand and object trajectories.
For \emph{pull} and \emph{push} actions, the metric requires that object is moved in the correct direction by at least \SI{50}{\milli\metre}.
During the object motion, the hand has to be oriented in the direction of the motion for push and in a reversed direction for pull actions (see Fig.~\ref{fig:metrics}).
Note that we also require the object stay upright (not fall) during the pull or push actions.
For \emph{pick something up} (not visualized), the metric checks that the object is held at least \SI{10}{\milli\metre} above the ground by the gripper.
Finally, for the two object actions (\emph{put}), we consider the final state of the objects.
The final position of the manipulated object should be within a \ang{120} arc around the static object, and within \SI{500}{\milli\metre} radius from the center of the static object.
The proposed metric is binary, and evaluates whether the policy executes the action for a given starting hand position and object size.

\FloatBarrier\section{Additional Results: Real2Sim}
\label{sec:app:exp-real2sim}

Quantitative results for different ablation methods for Real2Sim were presented as summaries in the main paper.
We elaborate on those in Table~\ref{tab:bench_states}, showing the fraction of correctly extracted trajectories for each action and method.
Recall, we analyze performance based on three methods depending on whether automatic or ground-truth was used for performing spatio-temporal localization:
A. ground-truth object/hand boxes and ground-truth action phase localization;
B. ground-truth boxes with predicted action phase; and
C. predicted boxes with predicted action phase.
We observe that method B works best, possibly due to the trajectory being estimated for the complete video followed by truncation of states using predicted action phase localization.
In contrast, for method A, we first truncate the video and then perform state estimation.

\begin{table}[h]
\centering
\tabcolsep=1.2mm
\begin{tabular}{ccccccccccc}
\toprule
\makecell{Method} 
& \makecell{Pull left\\ to right}
& \makecell{Pull right\\ to left}
& \makecell{Push left\\ to right}
& \makecell{Push right\\ to left }
& \makecell{Pick \\ up}
& \makecell{Put \\ behind}
& \makecell{Put in \\ front of}
& \makecell{Put \\ next to}
& \makecell{Put \\ onto}
& \textbf{Total} \\
\midrule
A & 6/6 & 4/6 & 5/6 & 6/6 & 2/6 & 2/6 & 5/6 & 4/6 & 2/6 & 36/54 \\
B & 6/6 & 5/6 & 6/6 & 6/6 & 3/6 & 2/6 & 5/6 & 6/6 & 0/6 & 39/54 \\
C & 5/6 & 4/6 & 5/6 & 6/6 & 2/6 & 1/6 & 4/6 & 6/6 & 0/6 & 33/54 \\
\bottomrule
\end{tabular}
\caption{Benchmarking states.}
\label{tab:bench_states}
\end{table}

\begin{figure}
\centering
\includegraphics[width=\linewidth]{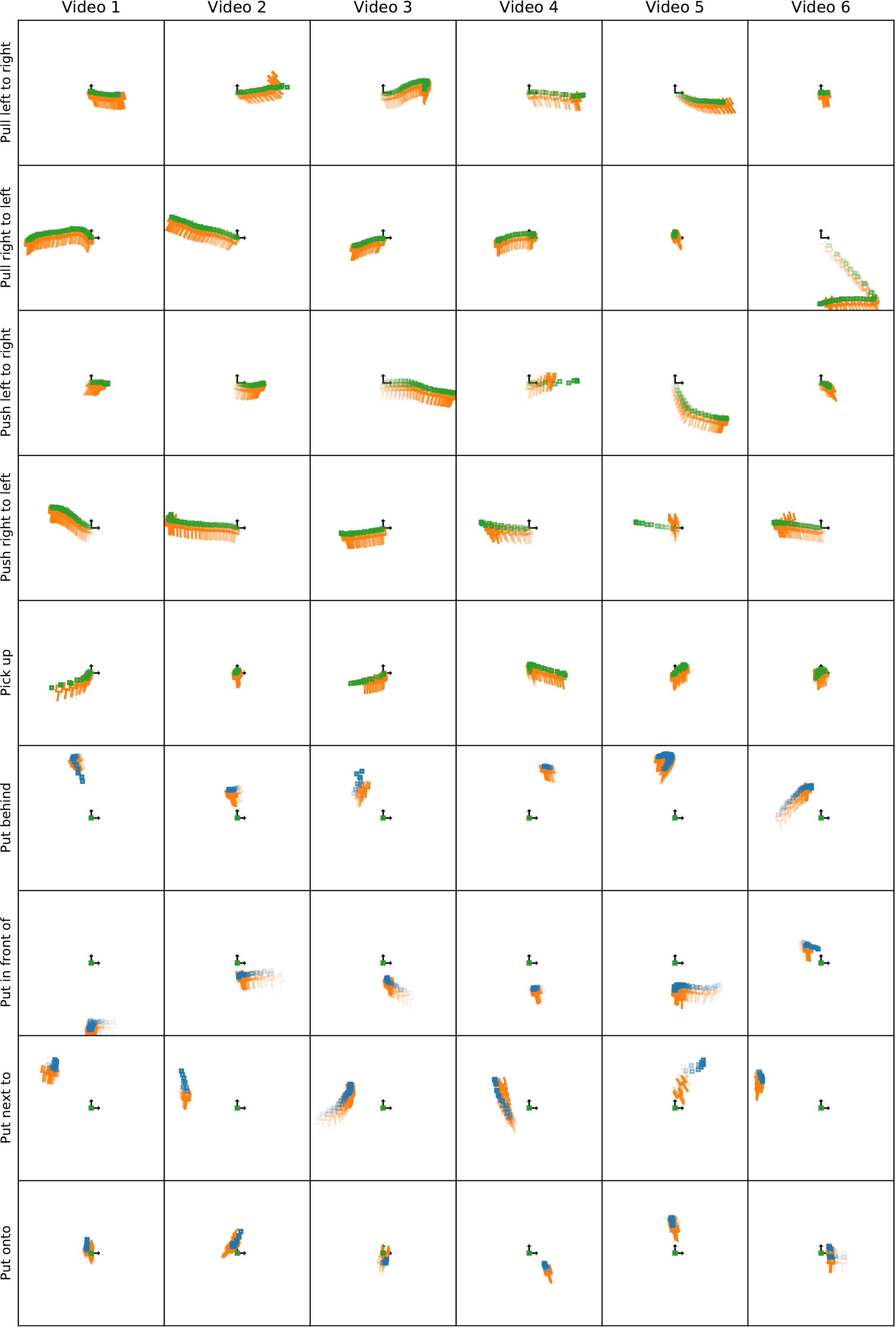}
\caption{Top view for reconstructed states of hand (orange) and objects (green, blue) in time using method C.
Transparency indicates temporal progression, more transparent means older in time.
}
\label{fig:top_view_s5}
\end{figure}

\textbf{Visualization.}
The metrics provide a limited understanding of the state estimation process (see Sec.~\ref{subsec:eval:ablation}).
We visualize the estimated hand and object trajectories from a top-view perspective in Fig.~\ref{fig:top_view_s5}.
To obtain better insights, we visualize additional results as GIFs in an HTML page~\texttt{real2sim.html} showing all 6 videos for each of the 9 actions.
In each row of the table, we show the full video, followed by segmentation masks $m^h_t, m^{o_i}_t$ and corresponding 3D state renderings via the neural renderer $r^h_t, r^{o_i}_t$.
All video ids correspond to the original ids in the Something-Something dataset.
The estimations from all three methods are shown as three columns of the table.

\FloatBarrier\section{Policy Training Results}
\label{sec:app:exp-policy}

We present additional results and analysis for training a policy with different settings.

\subsection{Curriculum learning}
\begin{figure}[h]
\centering
\includegraphics[width=\linewidth]{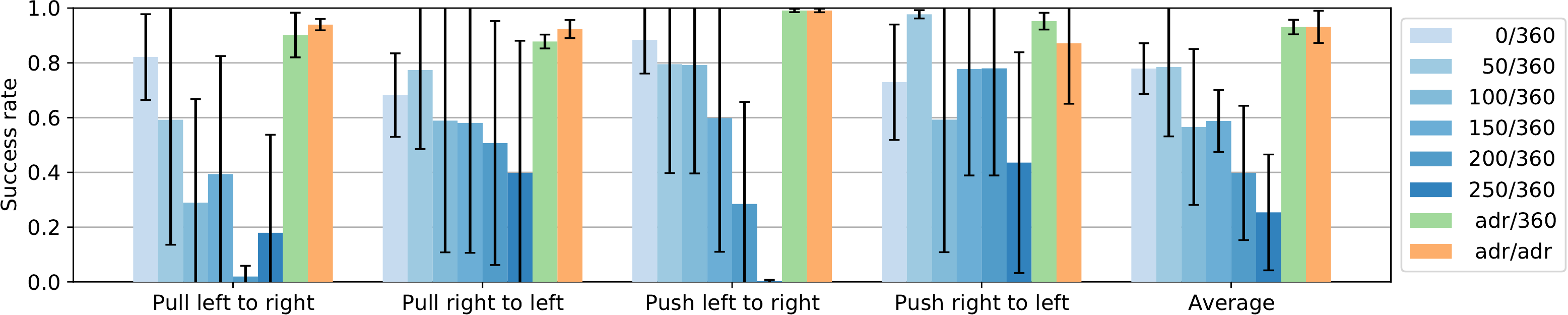}
\caption{
Success rate for multi-video policies trained with different randomization strategies using state trajectories from method A, and evaluated on the \textit{easy} benchmark.
\textbf{Legend}: the first number represents variability in the 3D position of the gripper, $\sigma$ in mm;
the second corresponds to the range of hand orientation in degrees.
Five policies are trained for each parameter and we report mean and standard deviation (error bar in the plot).
}
\label{fig:curriculum_easy}
\end{figure}

In the main paper, we looked at how curriculum learning with Automatic Domain Randomization (ADR) helped our agent to learn a better policy.
While the main paper presented results on the \emph{hard} benchmark (see Fig.~\ref{fig:curriculum}), here, we present additional results on the \emph{easy} benchmark in Fig.~\ref{fig:curriculum_easy}.
In particular, we observe the same trends.
The blue bars in Fig.~\ref{fig:curriculum_easy} show that the performance drops as $\sigma$ increases indicating that learning with a randomized gripper pose is challenging.
Our strategy to use ADR for both the 3D position and orientation (Fig.~\ref{fig:curriculum_easy} orange) achieves the best results and is used in all other experiments.

\FloatBarrier\subsection{Impact of GT in Real2Sim setups}
\label{sec:app:exp-policy:real2sim}
We present additional analysis of the difference in performance due to the use of GT bounding boxes or GT action phase.
Recall that these are referred to as method A, B, and C in the main paper.
On average, Fig.~\ref{fig:ablation} shows that the estimated vision works slightly worse than ground-truth as expected.
Actions \emph{put behind} and \emph{put onto} follow this trend.
For \emph{put next to}, the performance is within error bars.
By looking at \texttt{real2sim.html} (supplementary), we see that there are a few demonstrations where Real2Sim could be improved with better object segmentation and tracking.
We think these individual instances are outliers and are not representative of estimated vision working better, even for \emph{put in front of}. 
Referring to \texttt{real2sim.html} method A, we see that video~5 is challenging due to the elongated scissors, while video~6 fails as the object is placed too far for \emph{put next to}.
For \emph{put in front of}, video~1 places the object too far, and video~2 fails probably due to the GT box extending beyond the object mask.

\paragraph{Big performance gap for \emph{put onto}.}
We observe that method A achieves 60\% success rate for \emph{put onto}, while methods B and C are close to 0\%.
This can be explained by analyzing the GIFs for video~2 and video~4 in \texttt{real2sim.html}, where states estimated by method A are better than method C.
Quantitatively, method A gets 2/6 correct, while method C gets none (Table~\ref{tab:bench_states}).
RL is able to use these 2 trajectories to learn the task. Having ground-truth 2D boxes alleviates significant confusion while tracking the two objects.
Additionally, the big gap is also likely due to the binary nature of the metric that requires balancing objects: it either works, or the object falls, there is no half-way success.

\FloatBarrier\subsection{Learning from single vs. multiple videos.}
\label{sec:app:exp-policy:single-vs-multi}

Similar to Table~\ref{table:single_vs_multi_sets5hard} of the main paper that shows the performance of learning from single vs. multiple videos on the \emph{hard} benchmark with states obtained from method C, we present quantitative performance in a few more different settings.

In particular, Table~\ref{table:single_vs_multi_sets2easy} shows the performance on the \emph{easy} benchmark when using states estimated from method A.
Our agent is able to learn to perform most actions, often with success rates greater than 90\% (5/9 actions).
Table~\ref{table:single_vs_multi_sets2hard} presents the results on using the same states, but on the \emph{hard} benchmark.
We notice a drop in performance for all actions, particularly for starting positions of the hand being away from the camera.
Finally, Table~\ref{table:single_vs_multi_sets5easy} showcases the performance when using states estimated from method C on the \emph{easy} benchmark.

Overall, we observe that our proposed multi-video learning method outperforms learning from single videos, as some demonstrations may be noisy.
In addition, using sum (see Eq.~\eqref{eq:return_multivideo}) instead of max~\cite{peng2018deepMimic} shows consistent performance improvements across all experiments.

\begin{table}[h]
\centering
\tabcolsep=1.2mm
\begin{tabular}{lcccccc||ccc}
\toprule
Action &  vid 1 & vid 2 & vid 3 & vid 4 & vid 5 & vid 6 & average & proposed & baseline \\ 
\midrule
Pull left to right & \cellcolor{gray!0} 97 & \cellcolor{gray!0} 89 & \cellcolor{gray!0} 98 & \cellcolor{gray!0} 95 & \cellcolor{gray!0} 83 & \cellcolor{gray!119} 0 & \cellcolor{gray!3} 77 & \cellcolor{gray!0} 93 & \cellcolor{gray!39} 53 \\ 
Pull right to left & \cellcolor{gray!33} 57 & \cellcolor{gray!22} 64 & \cellcolor{gray!0} 94 & \cellcolor{gray!119} 0 & \cellcolor{gray!0} 94 & \cellcolor{gray!120} 0 & \cellcolor{gray!42} 51 & \cellcolor{gray!0} 91 & \cellcolor{gray!0} 95 \\ 
Push left to right & \cellcolor{gray!0} 99 & \cellcolor{gray!0} 99 & \cellcolor{gray!26} 62 & \cellcolor{gray!0} 98 & \cellcolor{gray!0} 98 & \cellcolor{gray!120} 0 & \cellcolor{gray!5} 76 & \cellcolor{gray!0} 99 & \cellcolor{gray!69} 33 \\ 
Push right to left & \cellcolor{gray!0} 98 & \cellcolor{gray!0} 84 & \cellcolor{gray!0} 99 & \cellcolor{gray!119} 0 & \cellcolor{gray!0} 99 & \cellcolor{gray!0} 99 & \cellcolor{gray!0} 80 & \cellcolor{gray!0} 92 & \cellcolor{gray!0} 93 \\ 
Pick up & \cellcolor{gray!0} 83 & \cellcolor{gray!54} 43 & \cellcolor{gray!23} 64 & \cellcolor{gray!36} 55 & \cellcolor{gray!0} 97 & \cellcolor{gray!119} 0 & \cellcolor{gray!33} 57 & \cellcolor{gray!0} 88 & \cellcolor{gray!45} 49 \\ 
Put behind & \cellcolor{gray!0} 99 & \cellcolor{gray!0} 99 & \cellcolor{gray!0} 94 & \cellcolor{gray!112} 5 & \cellcolor{gray!0} 89 & \cellcolor{gray!119} 0 & \cellcolor{gray!23} 64 & \cellcolor{gray!0} 95 & \cellcolor{gray!0} 95 \\ 
Put in front of & \cellcolor{gray!7} 75 & \cellcolor{gray!89} 20 & \cellcolor{gray!28} 60 & \cellcolor{gray!3} 78 & \cellcolor{gray!1} 79 & \cellcolor{gray!8} 74 & \cellcolor{gray!23} 64 & \cellcolor{gray!0} 79 & \cellcolor{gray!5} 76 \\ 
Put next to & \cellcolor{gray!0} 98 & \cellcolor{gray!11} 72 & \cellcolor{gray!30} 59 & \cellcolor{gray!0} 98 & \cellcolor{gray!52} 45 & \cellcolor{gray!15} 69 & \cellcolor{gray!8} 74 & \cellcolor{gray!18} 67 & \cellcolor{gray!6} 75 \\ 
Put onto & \cellcolor{gray!120} 0 & \cellcolor{gray!0} 89 & \cellcolor{gray!120} 0 & \cellcolor{gray!2} 78 & \cellcolor{gray!120} 0 & \cellcolor{gray!118} 1 & \cellcolor{gray!77} 28 & \cellcolor{gray!10} 73 & \cellcolor{gray!119} 0 \\ 
\bottomrule 
\end{tabular} 
\caption{
The success rate (in \%) for single and multi-video policies and for the baseline~\cite{peng2018deepMimic}.
Single video performances are averaged in the ``average'' column.
Results are on the \textit{easy} benchmark with states from method~A. 
} 
\label{table:single_vs_multi_sets2easy} 
\end{table}

\begin{table}[h]
\centering
\tabcolsep=1.2mm
\begin{tabular}{lcccccc||ccc}
\toprule
Action &  vid 1 & vid 2 & vid 3 & vid 4 & vid 5 & vid 6 & average & proposed & baseline \\ 
\midrule
Pull left to right & \cellcolor{gray!12} 71 & \cellcolor{gray!27} 61 & \cellcolor{gray!5} 76 & \cellcolor{gray!21} 65 & \cellcolor{gray!25} 62 & \cellcolor{gray!117} 1 & \cellcolor{gray!34} 56 & \cellcolor{gray!25} 62 & \cellcolor{gray!67} 35 \\ 
Pull right to left & \cellcolor{gray!60} 39 & \cellcolor{gray!49} 47 & \cellcolor{gray!9} 74 & \cellcolor{gray!115} 2 & \cellcolor{gray!23} 64 & \cellcolor{gray!120} 0 & \cellcolor{gray!63} 37 & \cellcolor{gray!29} 60 & \cellcolor{gray!5} 76 \\ 
Push left to right & \cellcolor{gray!0} 90 & \cellcolor{gray!11} 72 & \cellcolor{gray!73} 31 & \cellcolor{gray!0} 86 & \cellcolor{gray!0} 86 & \cellcolor{gray!120} 0 & \cellcolor{gray!28} 61 & \cellcolor{gray!0} 87 & \cellcolor{gray!78} 27 \\ 
Push right to left & \cellcolor{gray!0} 84 & \cellcolor{gray!30} 59 & \cellcolor{gray!0} 90 & \cellcolor{gray!114} 3 & \cellcolor{gray!0} 84 & \cellcolor{gray!0} 89 & \cellcolor{gray!17} 68 & \cellcolor{gray!0} 81 & \cellcolor{gray!1} 79 \\ 
Pick up & \cellcolor{gray!63} 37 & \cellcolor{gray!66} 36 & \cellcolor{gray!51} 45 & \cellcolor{gray!52} 45 & \cellcolor{gray!33} 57 & \cellcolor{gray!108} 7 & \cellcolor{gray!62} 38 & \cellcolor{gray!34} 57 & \cellcolor{gray!64} 36 \\ 
Put behind & \cellcolor{gray!0} 91 & \cellcolor{gray!0} 89 & \cellcolor{gray!0} 89 & \cellcolor{gray!72} 31 & \cellcolor{gray!0} 90 & \cellcolor{gray!119} 0 & \cellcolor{gray!21} 65 & \cellcolor{gray!0} 86 & \cellcolor{gray!0} 88 \\ 
Put in front of & \cellcolor{gray!52} 45 & \cellcolor{gray!97} 15 & \cellcolor{gray!58} 41 & \cellcolor{gray!40} 52 & \cellcolor{gray!31} 59 & \cellcolor{gray!36} 55 & \cellcolor{gray!52} 44 & \cellcolor{gray!35} 56 & \cellcolor{gray!42} 51 \\ 
Put next to & \cellcolor{gray!0} 88 & \cellcolor{gray!6} 76 & \cellcolor{gray!22} 65 & \cellcolor{gray!0} 94 & \cellcolor{gray!51} 45 & \cellcolor{gray!37} 55 & \cellcolor{gray!13} 70 & \cellcolor{gray!22} 65 & \cellcolor{gray!11} 72 \\ 
Put onto & \cellcolor{gray!120} 0 & \cellcolor{gray!21} 65 & \cellcolor{gray!119} 0 & \cellcolor{gray!32} 58 & \cellcolor{gray!120} 0 & \cellcolor{gray!119} 0 & \cellcolor{gray!88} 20 & \cellcolor{gray!33} 57 & \cellcolor{gray!118} 0 \\ 
\bottomrule 
\end{tabular} 
\caption{
The success rate (in \%) for single and multi-video policies and for the baseline~\cite{peng2018deepMimic}.
Single video performances are averaged in the ``average'' column.
Results are on the \textit{hard} benchmark with states from method~A.
} 
\label{table:single_vs_multi_sets2hard} 
\end{table}

\begin{table}[h]
\centering
\tabcolsep=1.2mm
\begin{tabular}{lcccccc||ccc}
\toprule
Action &  vid 1 & vid 2 & vid 3 & vid 4 & vid 5 & vid 6 & average & proposed & baseline \\ 
\midrule
Pull left to right & \cellcolor{gray!0} 99 & \cellcolor{gray!0} 88 & \cellcolor{gray!0} 88 & \cellcolor{gray!33} 57 & \cellcolor{gray!1} 79 & \cellcolor{gray!120} 0 & \cellcolor{gray!16} 68 & \cellcolor{gray!0} 93 & \cellcolor{gray!120} 0 \\ 
Pull right to left & \cellcolor{gray!0} 90 & \cellcolor{gray!0} 81 & \cellcolor{gray!0} 83 & \cellcolor{gray!90} 19 & \cellcolor{gray!120} 0 & \cellcolor{gray!120} 0 & \cellcolor{gray!51} 45 & \cellcolor{gray!26} 62 & \cellcolor{gray!119} 0 \\ 
Push left to right & \cellcolor{gray!0} 98 & \cellcolor{gray!0} 99 & \cellcolor{gray!0} 93 & \cellcolor{gray!8} 74 & \cellcolor{gray!36} 55 & \cellcolor{gray!120} 0 & \cellcolor{gray!14} 70 & \cellcolor{gray!0} 99 & \cellcolor{gray!120} 0 \\ 
Push right to left & \cellcolor{gray!0} 99 & \cellcolor{gray!0} 89 & \cellcolor{gray!0} 95 & \cellcolor{gray!0} 89 & \cellcolor{gray!0} 82 & \cellcolor{gray!0} 86 & \cellcolor{gray!0} 90 & \cellcolor{gray!0} 91 & \cellcolor{gray!0} 93 \\ 
Pick up & \cellcolor{gray!7} 75 & \cellcolor{gray!55} 43 & \cellcolor{gray!94} 17 & \cellcolor{gray!100} 12 & \cellcolor{gray!119} 0 & \cellcolor{gray!119} 0 & \cellcolor{gray!82} 24 & \cellcolor{gray!22} 64 & \cellcolor{gray!119} 0 \\ 
Put behind & \cellcolor{gray!0} 87 & \cellcolor{gray!0} 82 & \cellcolor{gray!70} 33 & \cellcolor{gray!88} 21 & \cellcolor{gray!112} 5 & \cellcolor{gray!113} 4 & \cellcolor{gray!61} 38 & \cellcolor{gray!0} 87 & \cellcolor{gray!0} 79 \\ 
Put in front of & \cellcolor{gray!0} 99 & \cellcolor{gray!0} 88 & \cellcolor{gray!34} 56 & \cellcolor{gray!5} 76 & \cellcolor{gray!27} 61 & \cellcolor{gray!73} 31 & \cellcolor{gray!16} 68 & \cellcolor{gray!0} 92 & \cellcolor{gray!0} 84 \\ 
Put next to & \cellcolor{gray!7} 74 & \cellcolor{gray!13} 71 & \cellcolor{gray!30} 59 & \cellcolor{gray!15} 69 & \cellcolor{gray!45} 50 & \cellcolor{gray!48} 47 & \cellcolor{gray!26} 62 & \cellcolor{gray!39} 53 & \cellcolor{gray!3} 77 \\ 
Put onto & \cellcolor{gray!82} 25 & \cellcolor{gray!115} 3 & \cellcolor{gray!115} 3 & \cellcolor{gray!117} 1 & \cellcolor{gray!120} 0 & \cellcolor{gray!118} 0 & \cellcolor{gray!111} 5 & \cellcolor{gray!117} 1 & \cellcolor{gray!118} 1 \\ 
\bottomrule 
\end{tabular} 
\caption{
The success rate (in \%) for single and multi-video policies and for the baseline~\cite{peng2018deepMimic}.
Single video performances are averaged in the ``average'' column.
Results are on the \textit{easy} benchmark with states from method~C. 
} 
\label{table:single_vs_multi_sets5easy} 
\end{table}

\textbf{Visualization.}
Similar to the results for Real2Sim, we create another HTML page, \texttt{policies.html} showing results of trained policies.
For each action, we present one successful and one failure example in the simulator, and the result of transferring the policy to a real robot.

\end{document}